\documentclass[nohyperref]{article}

\usepackage{microtype}
\usepackage{graphicx}
\usepackage{subfigure}
\usepackage{booktabs} 
\usepackage{multirow}
\usepackage{subfigure}

\usepackage{hyperref}
\usepackage{enumitem}
\usepackage{xurl}
\setlist{nosep}

\usepackage[accepted]{icml2023}

\usepackage{amsmath}
\usepackage{amssymb}
\usepackage{mathtools}
\usepackage{amsthm}

\usepackage[capitalize,noabbrev]{cleveref}

\theoremstyle{plain}

\theoremstyle{definition}

\theoremstyle{remark}

\usepackage[textsize=tiny]{todonotes}

\icmltitlerunning{Graph Reinforcement Learning for Network Control via Bi-Level Optimization}

\usepackage{amsmath}
\DeclareMathOperator*{\argmax}{arg\,max}
\DeclareMathOperator*{\argmin}{arg\,min}

\newcommand{\comm}{x}
\newcommand{\flow}{f}
\newcommand{\exch}{e}
\newcommand{\Exch}{E}
\newcommand{\money}{m}
\newcommand{\exweight}{w}

\newcommand{\nodes}{\mathcal{V}}
\newcommand{\storenodes}{\mathcal{V}_S}
\newcommand{\warehousenodes}{\mathcal{V}_W}
\newcommand{\edges}{\mathcal{E}}
\newcommand{\graph}{\mathcal{G}}
\newcommand{\order}{w}

\newcommand{\demand}{d}
\newcommand{\price}{p}
\newcommand{\capacity}{c}
\newcommand{\inventory}{q}
\newcommand{\bq}{\mathbf{q}}
\newcommand{\storagecost}{m^S}
\newcommand{\ordercost}{m^O}
\newcommand{\backordercost}{m^B}
\newcommand{\transportcost}{m^T}

\newcommand{\pol}{\pi}
\newcommand{\Pol}{\Pi}
\newcommand{\dist}{d}

\newcommand{\st}{s}
\newcommand{\stspace}{\mathcal{S}}
\newcommand{\ac}{a}
\newcommand{\acspace}{\mathcal{A}}
\newcommand{\dyn}{P}

\newcommand{\rew}{R}
\newcommand{\disc}{\gamma}

\newcommand{\R}{\mathbb{R}}
\newcommand{\E}{\mathbb{E}}

\newcommand{\ba}{\mathbf{a}}
\newcommand{\bx}{\mathbf{x}}
\newcommand{\be}{\mathbf{e}}
\newcommand{\bA}{\mathbf{A}}
\newcommand{\bD}{\mathbf{D}}
\newcommand{\bW}{\mathbf{W}}
\newcommand{\bI}{\mathbf{I}}

\usepackage[font=small,skip=4pt]{caption}
\begin{document}

\twocolumn[
\icmltitle{Graph Reinforcement Learning for Network Control \\
           via Bi-Level Optimization}



\icmlsetsymbol{equal}{*}

\begin{icmlauthorlist}
\icmlauthor{Daniele Gammelli}{stanford}
\icmlauthor{James Harrison}{google}
\icmlauthor{Kaidi Yang}{nus}
\icmlauthor{Marco Pavone}{stanford}
\icmlauthor{Filipe Rodrigues}{dtu}
\icmlauthor{Francisco C. Pereira}{dtu}
\end{icmlauthorlist}

\icmlaffiliation{stanford}{Stanford University}
\icmlaffiliation{google}{Google Research, Brain Team}
\icmlaffiliation{nus}{National University of Singapore}
\icmlaffiliation{dtu}{Technical University of Denmark}

\icmlcorrespondingauthor{Daniele Gammelli}{gammelli@stanford.edu}

\icmlkeywords{Network Optimization, Graph Neural Networks, Combinatorial Optimization}

\vskip 0.3in
]



\printAffiliationsAndNotice{}  

\begin{abstract}
Optimization problems over dynamic networks have been extensively studied and widely used in the past decades to formulate numerous real-world problems.
However, (1) traditional optimization-based approaches do not scale to large networks, and (2) the design of good heuristics or approximation algorithms often requires significant manual trial-and-error.
In this work, we argue that data-driven strategies can automate this process and learn efficient algorithms without compromising optimality.
To do so, we present network control problems through the lens of reinforcement learning and propose a graph network-based framework to handle a broad class of problems.
Instead of naively computing actions over high-dimensional graph elements, e.g., edges, we propose a bi-level formulation where we (1) specify a \textit{desired next state} via RL, and (2) solve a convex program to best achieve it, leading to drastically improved scalability and performance.
We further highlight a collection of desirable features to system designers, investigate design decisions, and present experiments on real-world control problems showing the utility, scalability, and flexibility of our framework.
\end{abstract}

\section{Introduction}
Many economically-critical real-world systems are well framed through the lens of control on graphs. 
For instance, the system-level coordination of power generation systems \cite{dommel1968optimal, huneault1991survey, bienstock2014chance}; road, rail, and air transportation systems \cite{WangSzetoEtAl2018, GammelliYangEtAl2021}; complex manufacturing systems, supply chain, and distribution networks \cite{sarimveis2008dynamic, bellamy2013network}; telecommunication networks \cite{jakobson1995real, flood1997telecommunication, popovskij2011control}; and many other systems can be cast as controlling flows of products, vehicles, or other quantities on graph-structured environments. 

A collection of highly effective solution strategies exist for versions of these problems. Some of the earliest applications of linear programming were network optimization problems \cite{dantzig1982reminiscences}, including examples such as maximum flow \cite{hillier1967introduction, sarimveis2008dynamic, ford1956maximal}. Within this context, handling multi-stage decision-making is typically addressed via time expansion techniques \cite{FordFulkerson1958, FordFulkerson1962}. However, despite their broad applicability, these approaches are limited in their ability to handle several classes of problems efficiently. Large-scale time-expanded networks may be prohibitively expensive, as are stochastic systems that require sampling realizations of random variables \cite{BirgeLouveaux2011, ShapiroDentchevaEtAl2014}. Moreover, nonlinearities may result in intractable optimization problems. 

\begin{figure*}[t]
\centering
    \includegraphics[width=0.9\textwidth]{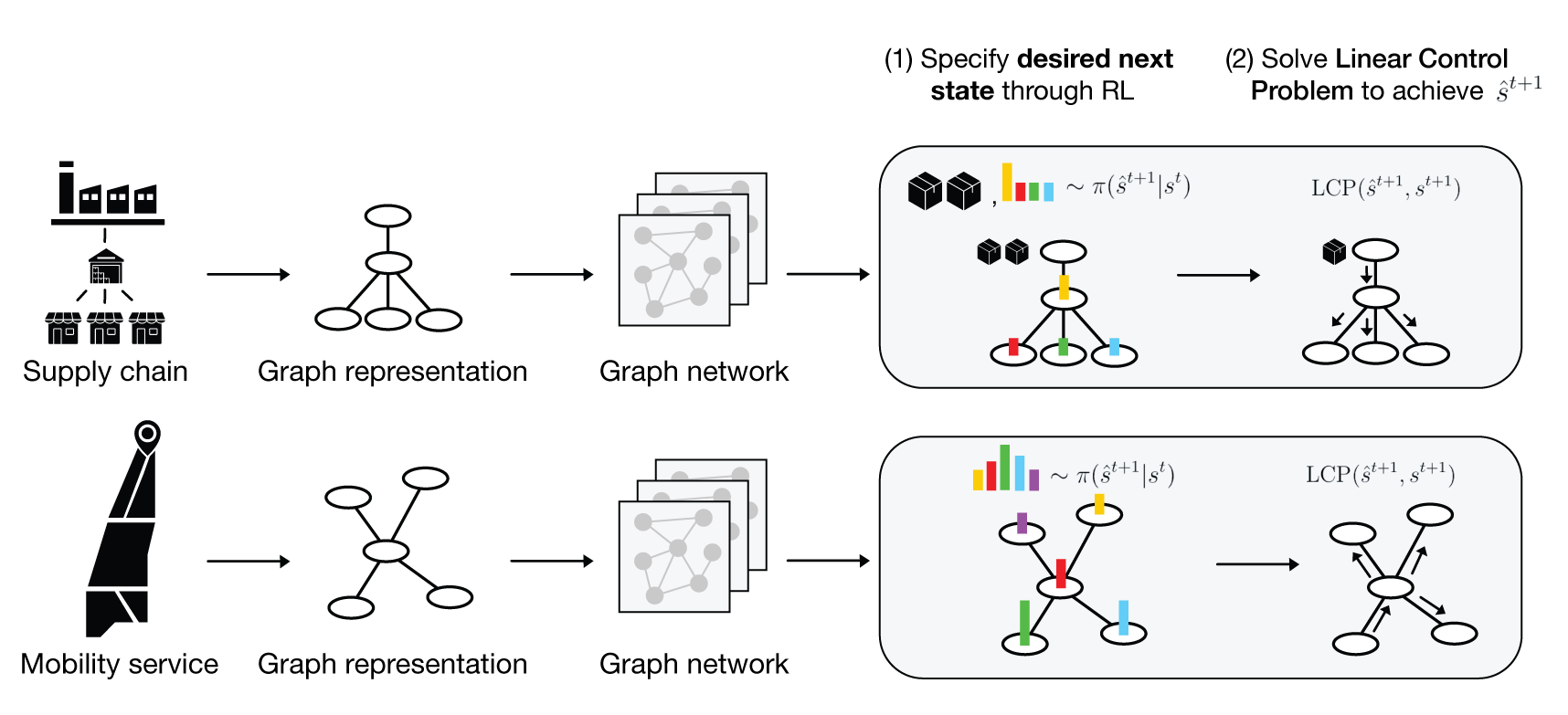}
    \caption{Many real-world systems (left) such as supply chain networks and mobility systems can be cast as controlling quantities within graph-structured environments (center-left). We present a framework that leverages graph networks (center) within a bi-level formulation. Instead of naively computing actions over graph elements, we first specify a \textit{desired next state} through RL (center-right), and then solve a convex program to compute the graph actions that can best achieve it (right).}
    \label{fig:graph_rl}
\end{figure*}

In this paper, we propose a strategy for simultaneously exploiting the tried-and-true optimization toolkit associated with network control problems while also handling the difficulties associated with stochastic, nonlinear, multi-stage decision-making. 
To do so, we present dynamic network problems through the lens of reinforcement learning and formalize a problem that is largely scattered across the control, management science, and optimization literature.
Specifically, we propose a learning-based framework to handle a broad class of network problems by exploiting the main strengths of graph representation learning, reinforcement learning, and classical operations research tools (Figure \ref{fig:graph_rl}). 

The contributions of this paper are threefold\footnote{Code available at: \url{https://github.com/DanieleGammelli/graph-rl-for-network-optimization}}:
\begin{itemize}
    \item We present a graph network-based bi-level, RL approach that leverages the specific strengths of direct optimization and reinforcement learning. 
    \item We investigate architectural components and design decisions within our framework, such as the choice of graph aggregation function, action parameterization, how exploration should be achieved, and their impact on system performance. 
    \item We show that our approach is highly performant, scalable, and robust to changes in operating conditions and network topologies, both on artificial test problems, as well as real-world problems, such as supply chain inventory control and dynamic vehicle routing. Crucially, we show that our approach outperforms classical optimization-based approaches, domain-specific heuristics, and pure end-to-end reinforcement learning. 
\end{itemize}

\section{Related Work}
Many real-world network control problems rely heavily on convex optimization \cite{BoydVandenberghe2004, hillier1967introduction}. This is often due to the relative simplicity of constraints and cost functions; for example, capacity constraints on edges may be written as simple linear combinations of flow values, and costs are linear in quantities due to the linearity of prices. In particular, linear programming (as well as specialized versions thereof) is fundamental in problems such as flow optimization, matching, cost minimization and optimal production, and many more.
While algorithmic improvements have made many convex problem formulations tractable and efficient to solve, these methods are still not able to handle (i) nonlinear dynamics, (ii) stochasticity, or (iii) the curse of dimensionality in time-expanded networks.
In this work, we aim to address these challenges by combining the strengths of direct optimization and reinforcement learning. 

Nonlinear dynamics typically requires linearization to yield a tractable optimization problem: either around a nominal trajectory, or iteratively during solution. While sequential convex optimization often yields an effective approximate solution, it is expensive and practically guaranteeing convergence while preserving efficiency may be difficult \cite{DinhDiehl2010}. Stochasticity may be handled in many ways: common strategies are distributional assumptions to achieve analytic tractability \cite{aastrom2012introduction}, building in sufficient buffer to correct via re-planning in the future \cite{powell2022reinforcement}, or sampling-based methods, often with fixed recourse \cite{ShapiroDentchevaEtAl2014}. Addressing the curse of dimensionality relies on limiting the amount of online optimization; typical approaches include limited-lookahead methods \cite{bertsekas2019reinforcement} or computing a parameterized policy via approximate dynamic programming or reinforcement learning \cite{Bertsekas1995, bertsekas1996neuro, SuttonBarto1998}. However, these policies may be strongly sub-optimal depending on representation capacity and state/action-space coverage. In contrast to these methods, we leverage the strong performance of optimization over short horizons (in which the impact of nonlinearity and stochasticity is typically limited) and exploit an RL-based heuristic for future returns which avoids the curse of dimensionality and the need to solve non-convex or sampled optimization problems. 

Our proposed approach results in a bi-level optimization problem.
Bi-level optimization---in which one optimization problem depends on the solution to another optimization problem, and is thus nested---has recently attracted substantial attention in machine learning, reinforcement learning, and control \cite{FinnAbbeelEtAl2017,harrison2018meta, AgrawalBarrattEtAl2019, AgrawalBarrattEtAl2019b, AmosKolter2017, landry2019bilevel, metz2019understanding}. Of particular relevance to our framework are methods that combine principled control strategies with learned components in a hierarchical way. Examples include using LQR control in the inner problem with learnable cost and dynamics \cite{tamar2017learning, amos2018differentiable, AgrawalBarrattEtAl2019b}, learning sampling distributions in planning and control \cite{ichter2018learning, power2022variational, amos2020differentiable}, or learning optimization strategies or goals for optimization-based control \cite{sacks2022learning, xiao2022learning, metz2019understanding, metz2022velo, LewSinghEtAl2022}. 

Numerous strategies for learning control with bi-level formulations have been proposed. A simple approach is to insert intermediate goals to train lower-level components, such as imitation \cite{ichter2018learning}. This approach is inherently limited by the choice of the intermediate objective; if this objective does not strongly correlate with the downstream task, learning could emphasize unnecessary elements or miss critical ones. An alternate strategy, which we take in this work, is directly optimizing through an inner controller, thus avoiding the problem of goal misspecification. A large body of work has focused on exploiting exact solutions to the gradient of (convex) optimization problems at fixed points \cite{amos2018differentiable, AgrawalBarrattEtAl2019b, donti2017task}. This allows direct backpropagation through optimization problems, allowing them to be used as a generic component in a differentiable computation graph (or neural network). Our approach leverages likelihood ratio gradients (equivalently, policy gradient), an alternate zeroth-order gradient estimator \cite{glynn1990likelihood}. This enables easy differentiation through lower-level optimization problems without the technical details required by fixed-point differentiation. 

\section{Problem Setting: Dynamic Network Control}
\label{sec:problem}

To outline our problem formulation, we first define the linear problem, which yields a classic convex problem formulation.
We will then define a nonlinear, dynamic, non-convex problem setting that better corresponds to real-world instances.
Much of the classical flow control literature and practice substitute the former linear problem for the latter nonlinear problem to yield tractable optimization problems \cite{li2007dcopf, ZhangRossiEtAl2016b, key1990distributed}.
Let us consider the control of $N_c$ \textit{commodities} on graphs - for example, vehicles in a transportation problem.
A graph $\mathcal{G} = \{\mathcal{V}, \mathcal{E}\}$ is defined as a set $\mathcal{V}$ of $N_v$ nodes, and a set $\mathcal{E}$ of $N_{\epsilon}$ ordered pairs of nodes $(i,j)$ called edges, each described by a travel time $t_{ij}$. 
We use $\mathcal{N}^{+}(i), \mathcal{N}^{-}(i) \subseteq \mathcal{V}$ for 
the set of nodes having edges pointing away from or toward node $i$, respectively.
We use $\comm^t_{i}(k) \in \R$ to denote the quantity of commodity $k$ at node $i$ and time $t$\footnote{We consider several reduced views over these quantities:
we write $\comm^t_{i} \in \R^{N_c}$ to denote the vector of all commodities, $\comm^t(k) \in \R^{N_v}$ to denote the vector of commodity $k$ at all nodes, and $\comm_i(k) \in \R^T$ to denote commodity $k$ at node $i$ for all times $t$. 
}.

\subsection{The Linear Network Control Problem}
\label{subsec:the_linear_control_problem}
Within the linear model, our commodity quantities evolve in time as 
\begin{equation}
    \comm_{i}^{t+1} = \comm_{i}^{t} + \flow_i^t + \exch_i^t, \quad \forall i \in \mathcal{V} \label{eq:con_flow}
\end{equation}
where $\flow_i^t$ denotes the change due to flow of commodities along edges and $\exch_i^t$ denotes the change due to exchange between commodities at the same graph node. We refer to this expression as the \textit{conservation of flow}. 
We also accrue money as
\begin{equation}
    \money^{t+1} = \money^{t} + \money_f^t + \money_e^t,
\end{equation}
where $\money_f^t, \money_e^t \in \R$ denote the money gained due to flows and exchanges respectively.
Our overall problem formulation will typically be to control \textbf{flows} and \textbf{exchanges} so as to maximize money over one or more steps subject to additional \textbf{constraints} such as, e.g., flow limitations through a particular edge. 

\textbf{Flows.} \hspace{2mm} 
We will denote flows along edge $(i,j)$ with $\flow_{ij}^t(k)$. From these flows, we have 
\begin{equation}
    \flow_i^t = \sum_{j\in\mathcal{N}^{-}(i)} \flow_{ji}^t - \sum_{j\in\mathcal{N}^{+}(i)} \flow_{ij}^t, \quad \forall i \in \mathcal{V} \label{eq:flow}
\end{equation}
which is the net flow (inflows minus outflows). As discussed, associated with each flow is a cost $\money_{ij}^t(k)$. Note that given this formulation, the total flow cost for all commodities can be written as $\money_{ij}^t \cdot \flow_{ij}^t = (\money_{ij}^t)^\top \flow_{ij}^t$. Thus, we can write the total flow cost at time $t$ as 
\begin{equation}
    \money_f^t = -\sum_{i \in \mathcal{V}}\left( \sum_{j\in\mathcal{N}^{-}(i)} \money_{ji}^t \cdot \flow_{ji}^t + \sum_{j\in\mathcal{N}^{+}(i)} \money_{ij}^t \cdot \flow_{ij}^t \right).\label{eq:money_flow}
\end{equation}

\textbf{Exchanges.} \hspace{2mm}
To define our exchange relations and their effect on commodity quantities and costs, we will write the effect that exchanges have on money for each node; we write this as $\money_i^t$. Thus, we have $\money_e^t = \sum_{i \in \mathcal{V}} \money_i^t$. We assume there are $N_e(i)$ exchange options at each node $i$. The exchange relation takes the form 
\begin{equation}
    \begin{bmatrix}
    \exch_i^t\\
    \money_i^t
    \end{bmatrix} = \Exch_i^t \exweight_i^t \label{eq:exch}
\end{equation}
where $\Exch_i^t \in \R^{(N_c+1) \times N_e(i)}$ is an exchange matrix and $\exweight \in \R^{N_e(i)}$ are the weights for each exchange. Each column in this exchange matrix denotes an (exogenous) exchange rate between commodities; for example, for $i$'th column $[-1, 1, 0.1]^\top$, one unit of commodity one is exchanged for one unit of commodity two plus $0.1$ units of money. Thus, the choice of exchange weights $\exweight_i^t$ uniquely determines exchanges $\exch_i^t$ and money change due to exchanges, $\money_e^t$.

\textbf{Convex constraints.} \hspace{2mm} 
We may impose additional convex constraints on the problem beyond the conservation of flow we have discussed so far. There are a few common examples that one may use in several applications. A common constraint is the non-negativity of commodity values, which we may express as
\begin{equation}
    \comm_i^t \geq 0, \quad \forall i, t. \label{eq:nonneg_comm}
\end{equation}
Note that this inequality is defined element-wise. We may also limit the flow of all commodities through a particular edge via
\begin{equation}
    \sum_{k=1}^{N_c} \flow_{ij}^t(k) \leq \overline{\flow}_{ij}^t ,\label{eq:flow_const}
\end{equation}
where this sum could also be weighted per commodity. These linear constraints are only a limited selection of some common examples and the particular choice of constraints is problem-specific. 

\subsection{The Nonlinear Dynamic Network Control Problem}
The previous subsection presented a linear, deterministic problem formulation that yields a convex optimization problem for the decision variables---the chosen flows and exchange weights. 
However, the formulation is limited by the assumption of linear, deterministic state transitions (among others), and is thus limited in its ability to represent typical real-world systems (please refer to Appendix \ref{appendix:dynamic_network_control} for a more complete treatment).
In this paper, we focus on solving the nonlinear problem (reflecting real, highly-general problem statements) via a bi-level optimization approach, wherein the linear problem (which has been shown to be extremely useful in practice) is used as an inner control primitive. 

\section{Methodology}
\label{sec:methodology}
In this section, we first introduce a Markov decision process (MDP) for our problem setting in Section \ref{subsec:dynamic_network_mdp}.
We further describe the bi-level formulation that is the primary contribution of this paper and provide insights on architectural considerations in Sections \ref{subsec:bilevel} and \ref{subsec:importance_of_gnns}, respectively.

\subsection{The Dynamic Network MDP}
\label{subsec:dynamic_network_mdp}
We consider a discounted MDP $\mathcal{M} = (\stspace, \acspace, \dyn, \rew, \disc)$. Here, $\st^t \in \stspace$ is the state and $\ac^t \in \acspace$ is the action, both at time $t$. The dynamics, $\dyn: \stspace \times \stspace \times \acspace \to [0,1]$ are probabilistic, with $\dyn(\st^{t+1}\mid\st^t,\ac^t)$ denoting a conditional distribution over $\st^{t+1}$. Finally, we use $\rew: \stspace \times \acspace \to \R$ to denote the reward function and $\disc \in (0,1]$ the discount factor.

\textbf{State and state space.} \hspace{2mm}
Real-world network control problems are typically partially-observed and many features of the world impact the state evolution. However, a small number of features are typically of primary importance, and the impact of the other partially-observed elements can be modeled as stochastic disturbances. 
Our formulation requires, at each timestep, the commodity values $\comm^t$. Furthermore, the constraint values are required, such as costs, exchange rates, flow capacities, etc. If the graph topology is time-varying, the connectivity at time $t$ is also critical. More precisely, the state elements that we have discussed so far are either properties of the graph nodes (commodity values) or of the edges (such as flow constraints). This difference is of critical importance in our graph neural network architecture.

Generally, the choice of state elements will depend on the information available to a system designer (what can be measured) and on the particular problem setting. Possible examples of further state elements include forecasts of prices, demand and supply, or constraints at future times. 

\textbf{Action and action space.} \hspace{2mm}
As discussed in Section \ref{sec:problem}, an action is defined as all flows and exchanges, $\ac^t = (\flow^t, \exweight^t)$.
In the following subsections, we accurately describe the action parametrization under the bi-level formulation.

\textbf{Dynamics.} \hspace{2mm}
The dynamics of the MDP, $\dyn$, describe the evolution of state elements. We split our discussion into two parts: the dynamics associated with commodity and non-commodity elements. 

The commodity dynamics are assumed to be reasonably well-modeled by the conservation of flow \eqref{eq:con_flow}, subject to the constraints; this forms the basis of the bi-level approach that we describe in the next subsection.

The non-commodity dynamics are assumed to be substantially more complex. For example, buying and selling prices may have a complex dependency on past sales, current demand, current supply (commodity values), as well as random exogenous factors. Thus, we place few assumptions on the evolution of non-commodity dynamics and assume that current values are measurable. 

\textbf{Reward.} \hspace{2mm}
We assume that our reward is the total discounted money earned over the problem duration. This results in a stage-wise reward function that corresponds to the money earned in that time period, or $\rew(\st^t,\ac^t) = \money_e^t + \money_f^t.$

\subsection{The Bi-Level Formulation}
\label{subsec:bilevel}
The previous subsection presented a general MDP formulation that represents a broad class of relevant network optimization problems.
The goal is to find a policy $\tilde{\pol}^* \in \tilde{\Pol}$ (where $\tilde{\Pol}$ is the space of realizable Markovian policies) such that $\tilde{\pol}^* \in \argmax_{\tilde{\pol} \in \tilde{\Pol}} \E_{\tau}\left[\sum_{t=0}^\infty \disc^t \rew(\st^t,\ac^t)\right]$, where $\tau = (\st^0, \ac^0, \st^1, \ac^1, \ldots)$ denotes the trajectory of states and actions. 
This formulation requires specifying a distribution over all flow/exchange actions, which may be an extremely large space. We instead consider a bi-level formulation
\begin{align}
    \pol^* \in & \argmax_{\pol \in \Pol} \E_{\tau} \left[\sum_{t=0}^\infty \disc^t \rew(\st^t,\ac^t)\right] \\ 
    & \text{s.t.} \,\,\ac^t = \text{LCP}(\hat{\st}^{t+1}, \st^t),
\end{align}
where we compute $\ac^t$ by replacing a single policy that maps from states to actions (i.e., $\st^t \rightarrow \ac^t$) with two nested policies, mapping from states to desired next states to actions (i.e., $\st^t \rightarrow \hat \st^{t+1} \rightarrow \ac^t$).
As a consequence of this formulation, the desired next state $\hat \st^{t+1}$ acts as an intermediate variable, thus avoiding the direct parametrization of an extremely large action space, e.g., flows over edges in a graph. 
This desired next state is then used in a linear control problem ($\text{LCP}(\cdot,\cdot)$), which leverages a (slightly modified) one-step version of the linear problem formulation of Section \ref{sec:problem} to map from desired next state to action. 
Thus, the resulting formulation is a bi-level optimization problem, whereby the policy $\tilde{\pol}$ is the composition of the policy $\pol(\hat{\st}^{t+1}\mid \st^t)$ and the solution to the linear control problem. 
Specifically, given a sample of $\hat{\st}^{t+1}$ from the stochastic policy, we select flow and exchange actions by solving
\begin{subequations}
\begin{align}
    \argmin_{\ac^t} \quad& \dist(\hat{\st}^{t+1}, {\st}^{t+1}) - \rew(\st^t, \ac^t) \label{eq:reb_obj}\\
    \rm{s.t.}  \quad \,\,\,\,\,
    & \text{Conservation of flow } \eqref{eq:con_flow}; \text{Net flow }\eqref{eq:flow}; \\
    & \text{Reward }\eqref{eq:money_flow}; \text{Exchange conditions }\eqref{eq:exch}; \\
    & \text{Other constraints, e.g.~} \eqref{eq:nonneg_comm} \text{ or } \eqref{eq:flow_const}
\end{align}\label{eq:reb}
\end{subequations}
where $d(\cdot, \cdot)$ is a convex metric which penalizes deviation from the desired next state. The resultant problem is convex and thus may be easily and inexpensively solved to choose actions $\ac^t$, even for very large problems. 
Please see Appendix \ref{appendix:subsec:rl_details}, \ref{appendix:discussion} for a broader discussion.

As is standard in reinforcement learning, we will aim to solve this problem via learning the policy from data. 
This may be in the form of online learning \cite{SuttonBarto1998} or via learning from offline data \cite{LevineEtAl2020}.
There are large bodies of work on both problems, and our presentation will generally aim to be as-agnostic-as-possible to the underlying reinforcement learning algorithm used. Of critical importance is the fact that the majority of reinforcement learning algorithms use likelihood ratio gradient estimation \cite{Williams1992}, which does not require path-wise backpropagation through the inner problem. 

We also note that our formulation assumes access to a model (the linear problem) that is a reasonable approximation of the true dynamics over short horizons. 
This short-term correspondence is central to our formulation: we exploit exact optimization when it is useful, and otherwise push the impacts of the nonlinearity over time to the learned policy. We assume this model is known in our experiments---which we feel is a reasonable assumption across the problem settings we investigate---but it could be learned from state transitions or as learnable parameters in policy learning.

\subsection{Architectural Considerations}
\label{subsec:importance_of_gnns}
After having introduced the problem formulation and a general framework to control graph-structured systems from experience, here and in Appendix \ref{appendix:subsec:network_architeture}, we broaden the discussion on specific algorithmic components.

\begin{table*}[t]
\centering
\caption{Percentage of oracle performance on different minimum cost flow scenarios.}
\footnotesize
\begin{tabular}{l c c c c c c}
    & Random & MLP-RL & GCN-RL & GAT-RL & MPNN-RL & Oracle \\
    \midrule 
    2-hops &  9.9\% $\pm$4.8\% & 60.2\% $\pm$2.1\% & 31.3\% $\pm$1.3\% & 22.9\% $\pm$1.1\% & \textbf{89.7\% $\mathbf{\pm0.9}$\%} & - \\[0.1ex]
    3-hops &   50.3\% $\pm$8.4\% & 53.8\% $\pm$1.6\% & 68.7\% $\pm$2.0\% & 62.4\% $\pm$1.9\% & \textbf{89.5\% $\mathbf{\pm1.1}$\%} & - \\[0.1ex]
    4-hops &   63.1\% $\pm$3.9\% & 67.8\% $\pm$2.5\% & 71.4\% $\pm$1.7\% & 68.2\% $\pm$2.3\% & \textbf{87.1\% $\mathbf{\pm1.2}$\%} & - \\[0.1ex]
    Dynamic travel time &  -23.4\% $\pm$4.3\% & -0.7\% $\pm$1.7\% & 18.7\% $\pm$2.0\% & 17.1\% $\pm$1.6\% & \textbf{99.1\% $\mathbf{\pm1.3}$\%} & - \\[0.1ex]
    Dynamic topology &  42.5\% $\pm$6.8\% & N/A & 53.4\% $\pm$2.8\% & 43.4\% $\pm$3.1\% & \textbf{83.9\% $\mathbf{\pm1.0}$\%} & - \\[0.1ex]
    Multi-commodity &  22.5\% $\pm$8.2\% & 41.7\% $\pm$3.2\% & 33.8\% $\pm$2.1\% & 33.0\% $\pm$1.7\% & \textbf{72.0\% $\mathbf{\pm1.6}$\%} & - \\[0.1ex]
    Capacity (Success Rate) & 62.6\% (82\%) & 62.7\% (82\%) & 65.2\% (87\%) & 62.9\% (80\%) & \textbf{89.8\%} \textbf{(87\%)} & - (88\%) \\[0.1ex]
    \bottomrule
    \end{tabular}%
\label{tab:experiments}%
\end{table*}

\textbf{Network architectures.} \hspace{2mm}
We argue that graph networks represent a natural choice for network optimization problems because of three main properties.
First, permutation invariance. 
Crucially, non-permutation invariant computations would consider each node ordering as fundamentally different and thus require an exponential number of input/output training examples before being able to generalize.
Second, locality of the operator. 
GNNs typically express a local parametric filter (e.g., convolution operator) which enables the same neural network to be applied to graphs of varying size and connectivity and achieve non-parametric expansibility.
This is a property of fundamental importance for many real-world graph control problems, which will be dynamic or frequently re-configured, and it is desirable to be able to use the same policy without re-training.
Lastly, alignment with the computations used for network optimization problems.
As shown in \cite{XuEtAl2020}, GNNs can better match the structure of many network optimization algorithms and are thus likely to achieve better performance.

\textbf{Action parametrization.} \hspace{2mm}
Let us consider the problem of controlling flows in a network.
We are interested in defining a desired next state $\hat \st^{t+1}$ that is ideally (i) lower dimensional, (ii) able to capture relevant aspects for control, and (iii) as-robust-as-possible to domain shifts.
At a high level, we achieve this by avoiding the direct parametrization of per-edge desired flow values and compute per-node desired inflow quantities. 
Concretely, given the total availability $M$ of commodity units in the graph, we define $\hat \st^{t+1} = \{\hat \inventory_i^{t+1}\}_{i \in \nodes}, \sum_i \hat \inventory_i^{t+1} = M$ as a desired per-node number of commodity units.
We do so by first determining $\tilde{\inventory}_{i}^{t+1} = \{\tilde{\inventory}_{i}^{t+1}\}_{i \in \nodes}$, where $\tilde{\inventory}_{i}^{t+1} \in [0,1]$ defines the percentage of currently available commodity units to be moved to node $i$ in time step $t$, and $\sum_{i \in \nodes} \tilde{\inventory}_{i}^{t+1} = 1$. 
We then use this to compute $\hat{\inventory}_i^{t+1}=\lfloor \tilde{\inventory}_{i}^{t+1} \cdot M\rfloor$ as the actual number of commodity units.
In practice, we achieve this by defining the intermediate policy as a Dirichlet distribution over nodes, i.e., $\pol(\hat \st^{t+1} | \st^t) =  \tilde{\inventory}^{t+1} \cdot M, \tilde{\inventory}^{t+1} \sim \text{Dir}(\tilde{\inventory}^{t+1} | \st^t)$.
Crucially, the representation of the desired next state via $\hat \inventory_i$ (i) is lower-dimensional as it only acts over nodes in the graph, (ii) uses a meaningful aggregated quantity to control flows, and (iii) is scale-invariant by construction as it acts on \textit{ratios} opposed to raw commodity quantities.
Additionally, for problems that require a generation of commodities (e.g., products in a supply chain), we define the desired next state via the exchange weights introduced in Eq \eqref{eq:exch}, $\hat \st^{t+1} = \{w_i^{t+1}\}_{i \in \nodes}, w_i^{t+1} \in \mathbb{N}^+$, with $w_i^{t+1}$ representing the number of commodity units to generate.
In practice, this can be achieved by defining the intermediate policy as a Gaussian distribution over nodes (followed by rounding), i.e., $\pol(\hat \st^{t+1} | \st^t) =  \text{round}(w^{t+1}), w^{t+1} \sim \mathcal{N}(w^{t+1} | \st^t)$.

\section{Experiments}
\label{sec:experiments}
In this section, we first consider an artificial \textit{minimum cost flow problem} as a simple graph control problem that illustrates the basic principles of our formulation and investigates architectural components (Section \ref{subsec:minimum_cost_flow}).
We further assess the versatility of our framework by applying it to two distinct real-world network problems: the \textit{supply chain inventory management} problem (Section \ref{subsec:supply_chain_inventory_management}) and the \textit{dynamic vehicle routing} problem (Section \ref{subsec:dynamic_vehicle_routing}).
Specifically, these problems represent two instantiations of economically-critical graph control problems where the task is to control flows of quantities (i.e., packages and vehicles, respectively), generate commodities (i.e., products within a supply chain), or both.

\textbf{Experimental design.} \hspace{2mm}
While the specific benchmarks will necessarily depend on the individual problem, in all real-world experiments, we will always compare against the following \textit{classes} of methods: (i) an \textit{Oracle} benchmark characterized as an MPC controller which has access to perfect information of all future states of the system and can thus plan for the perfect action sequence, (ii) a \textit{Domain-driven Heuristic}, i.e., algorithms which are generally accepted as go-to approaches for the types of problems we consider, and (iii) a \textit{Randomized} heuristic to quantify a reasonable lower-bound of performance within the environment.

\subsection{Minimum Cost Flow}
\label{subsec:minimum_cost_flow}
\begin{figure}[t]
      \centering
     \includegraphics[width=\columnwidth]{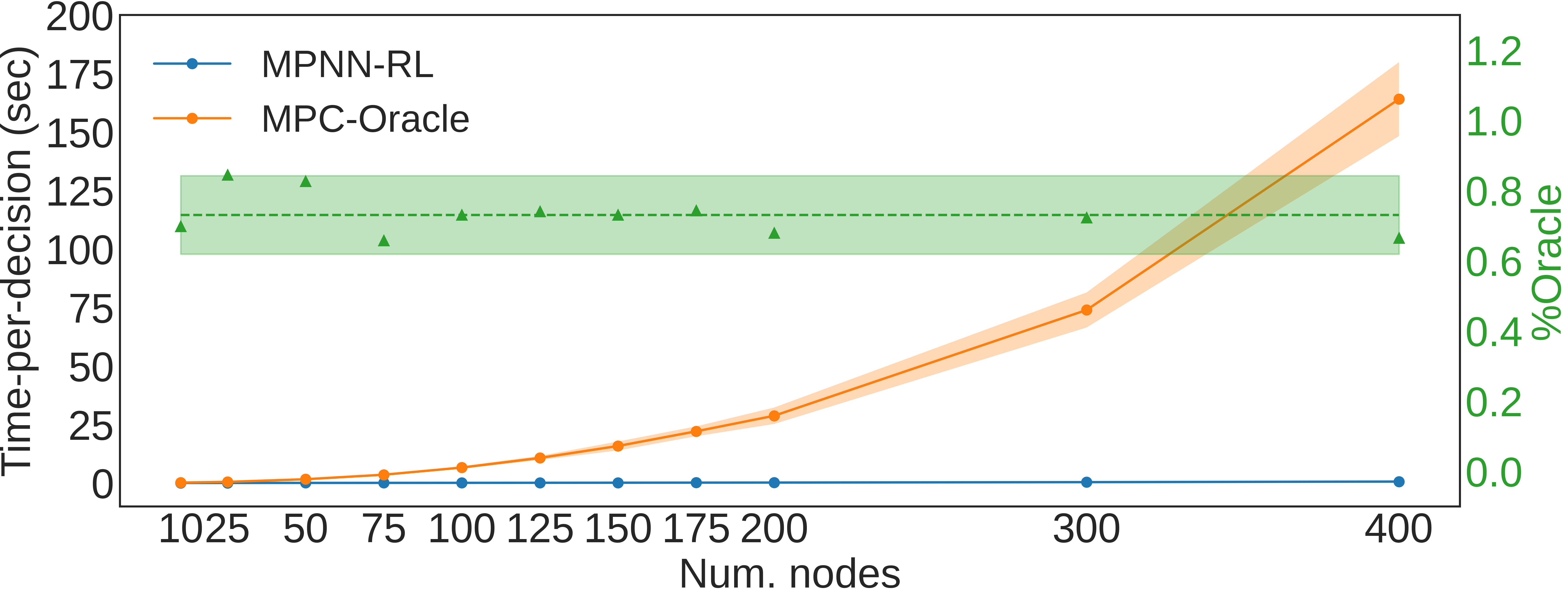}
      \caption{Comparison of computation times between learning-based (blue) and control-based (orange) approaches. Green triangles represent the percentage performance of our RL framework compared to the oracle model.}
      \label{fig:computational_analysis}
\end{figure}

Let us consider an artificial minimum cost flow problem where the goal is to control commodities from one or more source nodes to one or more sink nodes, in the minimum time possible.
We assess the capability of our formulation to handle several practically-relevant situations.
Specifically, we do so by comparing different versions of our method against an oracle benchmark to investigate the effect of different neural network architectures.
Results in Table \ref{tab:experiments} and in Appendix \ref{appendix:subsubsec:min_cost_flow_additional_results}, show how graph-RL approaches are able to achieve close-to-optimal performance in all proposed scenarios while greatly reducing the computation cost compared to traditional solutions (Figure \ref{fig:computational_analysis} and Appendix \ref{appendix:subsec:computational_efficiency})\footnote{All methods used the same computational CPU resources, namely a AMD Ryzen Threadripper 2950X (16-Core, 32 Thread, 40M Cache, 3.4 GHz base).}.
Among all formulations (please refer to Appendix \ref{appendix:subsec:minimum_cost_flow} for additional details), MPNN-RL is clearly the best performing architecture, achieving $86.7\%$ of oracle performance, on average.
As discussed in Section \ref{subsec:importance_of_gnns}, this highlights the importance of the algorithmic alignment \cite{XuEtAl2020} between the neural network architecture and the nature of the computations needed to solve the task.
Crucially, results show how our formulation is able to operate reliably within a broad set of situations, ranging from scenarios characterized by dynamic travel times (\textit{Dyn. travel time}), dynamic topologies, i.e., with nodes and edges that can be removed or added during an episode (\textit{Dyn. topology}), capacitated-networks (\textit{Capacity}) with different depth (\textit{2-hop}, \textit{3-hop}, \textit{4-hop}), and multi-commodity problems (\textit{Multi-commodity}).

\subsection{Supply Chain Inventory Management (SCIM)}
\label{subsec:supply_chain_inventory_management}
\begin{table*}[t]
\centering
\small
\caption{System performance on real-world SCIM experiments.}
\begin{tabular}{l c c c c c}
    & Avg. Prod. & S-type Policy & End-to-End RL (MLP/GNN) & Graph-RL (ours) & Oracle \\
    \midrule 
    1F 2S & -20,334 ($\pm$ 4,723) & -4,327 ($\pm$ 251) & -1,832 ($\pm$ 352) / -17 ($\pm$ 89) & \textbf{192 ($\pm$ 119)} & 852 ($\pm$ 152) \\ [0.1ex]
    \% Oracle & 0.0\% & 75.5\% & 87.3\% / 95.8\% & \textbf{96.8\%} & 100.0\% \\  [0.1ex]
     1F 3S  & -53,113 ($\pm$ 7,231) & -5,650 ($\pm$ 298) & -4,672 ($\pm$ 258) / -810 ($\pm$ 258) & \textbf{997 ($\pm$ 109)} & 3,249 ($\pm$ 102) \\ [0.1ex]
     \% Oracle & 0.0\% & 84.2\% & 85.9\% / 92.7\% & \textbf{96.0\%} & 100.0\% \\  [0.1ex]
     1F 10S & -114,151 ($\pm$ 4,611) & -14,327 ($\pm$ 365) & -587,887 ($\pm$ 5,255) / -568,374 ($\pm$ 5,255) & \textbf{890 ($\pm$ 288)} & 1,358 ($\pm$ 460) \\ [0.1ex]
     \% Oracle & 0.0\% & 86.4\% & N.A. /  N.A. & \textbf{99.5\%} & 100.0\% \\  [0.1ex]
    \bottomrule
    \end{tabular}%
\label{tab:real_world_experiments_scim}%
\end{table*}

In our first real-world experiment, we aim to optimize the performance of a supply chain inventory system.
Specifically, this describes the problem of ordering and shipping product inventory within a network of interconnected warehouses and stores in order to meet customer demand while simultaneously minimizing storage and transportation costs.
A supply chain system is naturally expressed via a graph $\graph = \{\nodes, \edges\}$, where $\nodes = \storenodes \cup \warehousenodes$ is the set of both store $\storenodes$ and warehouse $\warehousenodes$ nodes, and $\edges$ the set of edges connecting stores to warehouses.
Demand $\demand_i^t$ materializes in stores $i \in \storenodes$ at each period $t$. 
If inventory is available at the store, it is used to meet customer demand and sold at a price $\price$.
Unsatisfied orders are maintained over time and are represented as a negative stock (i.e., backorder).
At each time step, the warehouse orders additional units of inventory $\order_i$ from the manufacturers and stores available ones.
As commodities travel across the network, they are delayed by transportation times $t_{ij}$.
Both warehouses and storage facilities have limited storage capacities $\capacity_i$, such that the current inventory $\inventory_i$ cannot exceed it.
The system incurs a number of operations-related costs: storage costs $\storagecost_i$, production costs $\ordercost_i$, backorder costs $\backordercost_i$, transportation costs $\transportcost_{ij}$.

\textbf{SCIM Markov decision process.} \hspace{2mm}
To apply the methodologies introduced in Section \ref{sec:methodology}, we formulate the SCIM problem as an MDP characterized by the following elements (please refer to Appendix \ref{appendix:subsubsec:scim_mdp_details} for a formal definition):

\textit{Action space ($\acspace$):} we consider the problem of determining (1) the amount of additional inventory $\order_i$ to order from manufacturers in all warehouse nodes $i \in \warehousenodes$, and (2) the flow $\flow_{ij}$ of commodities to be shipped from warehouses to stores, such that $\ba^t = \{\order_i^t\}_{ i\in\warehousenodes} \cup \{\flow_{ij}^t\}_{(i,j) \in \edges}$.

\textit{Reward $\left(R(\st^t, \ac^t)\right)$:} we select the reward function in the MDP as the profit of the inventory manager, computed as the difference between sales revenues and costs. 

\textit{State space ($\stspace$):} the state space describes the current status of the supply network, via node and edge features.
Node features contain information on (i) current inventory, (ii) current and estimated demand, (iii) incoming flow, and (iv) incoming orders.
Edge features are characterized by (i) travel time $t_{ij}$, and (ii) transportation cost $\transportcost_{ij}$.

\textbf{Bi-Level formulation.} \hspace{2mm}
In what follows and in Appendix \ref{appendix:subsubsec:scim_lcp}, we illustrate a specific instantiation of our framework for the SCIM problem.
We define the desired outcome $\hat \st^{t+1}$ as being characterized by two elements: (i) the desired production in warehouse nodes $\hat \order_i^{t+1}, \forall i \in \warehousenodes$, and (ii) a desired inventory in store nodes $\hat \inventory_i^{t+1}, \forall i \in \storenodes$.
The LCP selects flow and production actions to best achieve $\hat \st^{t+1}$ via distance minimization between desired and actual inventory levels. 
The LCP is further defined by domain-related constraints, such as ensuring that the inventory in store and warehouse nodes does not exceed storage capacity and that shipped products are non-negative and upper bounded by inventory.

\textbf{Inventory management via graph control.} \hspace{2mm}
\begin{figure}[t]
    \centering
    \includegraphics[width=\columnwidth]{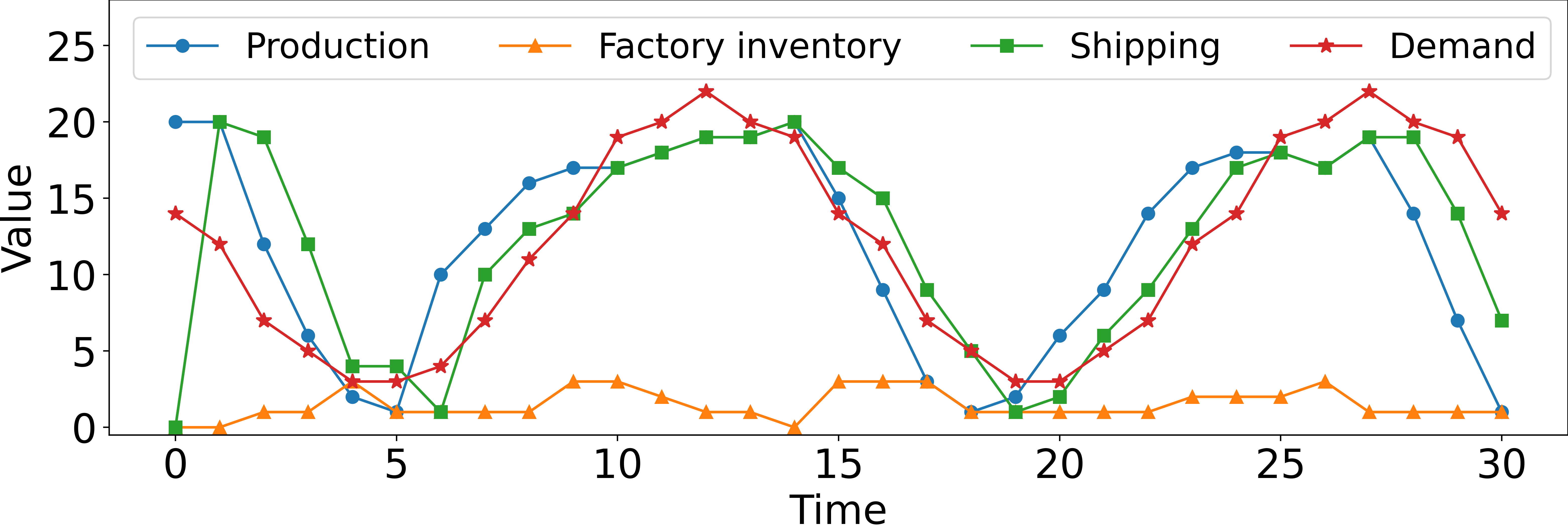}
    \caption{Aggregate behavior of the Graph-RL policies on a test episode for the 1F2S SCIM environment.}
    \label{fig:scim_experiments}
\end{figure}
For the SCIM problem, we define the domain-driven heuristic as a prototypical S-type (or ``order-up-to'') policy, which is generally accepted as an effective heuristic \cite{VanRoyEtAl1997}.
Appendix \ref{appendix:subsec:scim} provides further experimental details.

Concretely, we measure overall system performance on three different supply chain networks characterized by increasing complexity.
Results in Table \ref{tab:real_world_experiments_scim} show that our framework achieves close-to-optimal performance in all tasks. 
Specifically, Graph-RL achieves 96.8\% (1F2S), 96\% (1F3S), and 99.5\% (1F10S) of oracle performance.
Qualitatively, Figure \ref{fig:scim_experiments} highlights how Graph-RL learns to control the production and shipping policies to match consumer demand while maintaining low inventory storage.
More subtly, Figure \ref{fig:scim_experiments} shows how policies learned through Graph-RL manage to anticipate demand so that products are promptly available in stores by taking production and shipping time under consideration.
Results in Table \ref{tab:real_world_experiments_scim} also show how S-type policies, despite being explicitly fine-tuned for all tasks, are largely inefficient and thus incur unnecessary costs and revenue losses, resulting in a profit gap of approximately $15\%$ compared to Graph-RL, on average. 

\label{subsec:dynamic_vehicle_routing}
\begin{table*}[t]
\centering
\footnotesize
\caption{System performance on real-world DVR experiments.}
\begin{tabular}{l c c c c c}
    & Random & Evenly-balanced System & End-to-end RL & Graph-RL (ours) & Oracle \\
    \midrule 
    New York & -10,778 ($\pm$ 659) & 9,037 ($\pm$ 797) & -6,043 ($\pm$ 2,584) & \textbf{15,481 ($\pm$ 397)} & 16,867 ($\pm$ 547)\\ [0.1ex]
    \% Oracle & 0.0\% & 71.6\% & 17.2\% & \textbf{94.9\%} & 100.0\% \\  [0.1ex]
    Shenzhen & 19,406 ($\pm$ 1,894) & 29,826 ($\pm$ 706) & 18,889 ($\pm$ 1,207) & \textbf{36,918 ($\pm$ 616)} & 40,332 ($\pm$ 724)\\ [0.1ex] 
    \% Oracle & 0.0\% & 50.1\% & -0.02\% & \textbf{83.8\%} & 100.0\% \\  [0.1ex]
    \midrule
    Zero Shot NY$\rightarrow$SHE & - & - & 18,568 ($\pm$ 1,358) & 36,100 ($\pm$ 657) & - \\ [0.1ex]
    Zero Shot SHE$\rightarrow$NY & - & - & -4,083  ($\pm$ 1,278) & 14,495 ($\pm$ 426) & - \\ [0.1ex]
    \bottomrule
    \end{tabular}%
\label{tab:real_world_experiments_dvr}%
\end{table*}
\begin{figure*}
    \centering
    \subfigure[]{\includegraphics[width=0.33\textwidth]{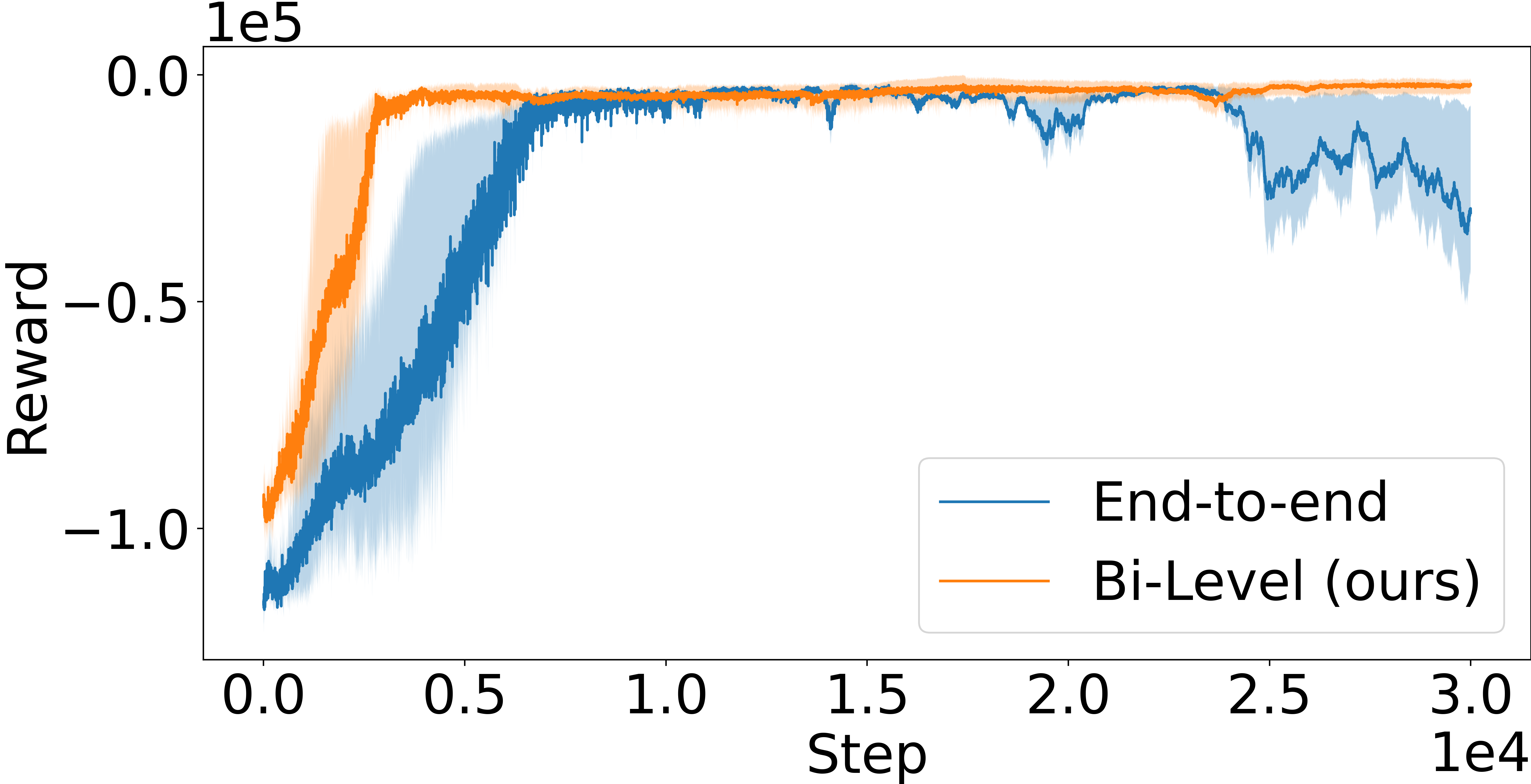}} 
    \subfigure[]{\includegraphics[width=0.33\textwidth]{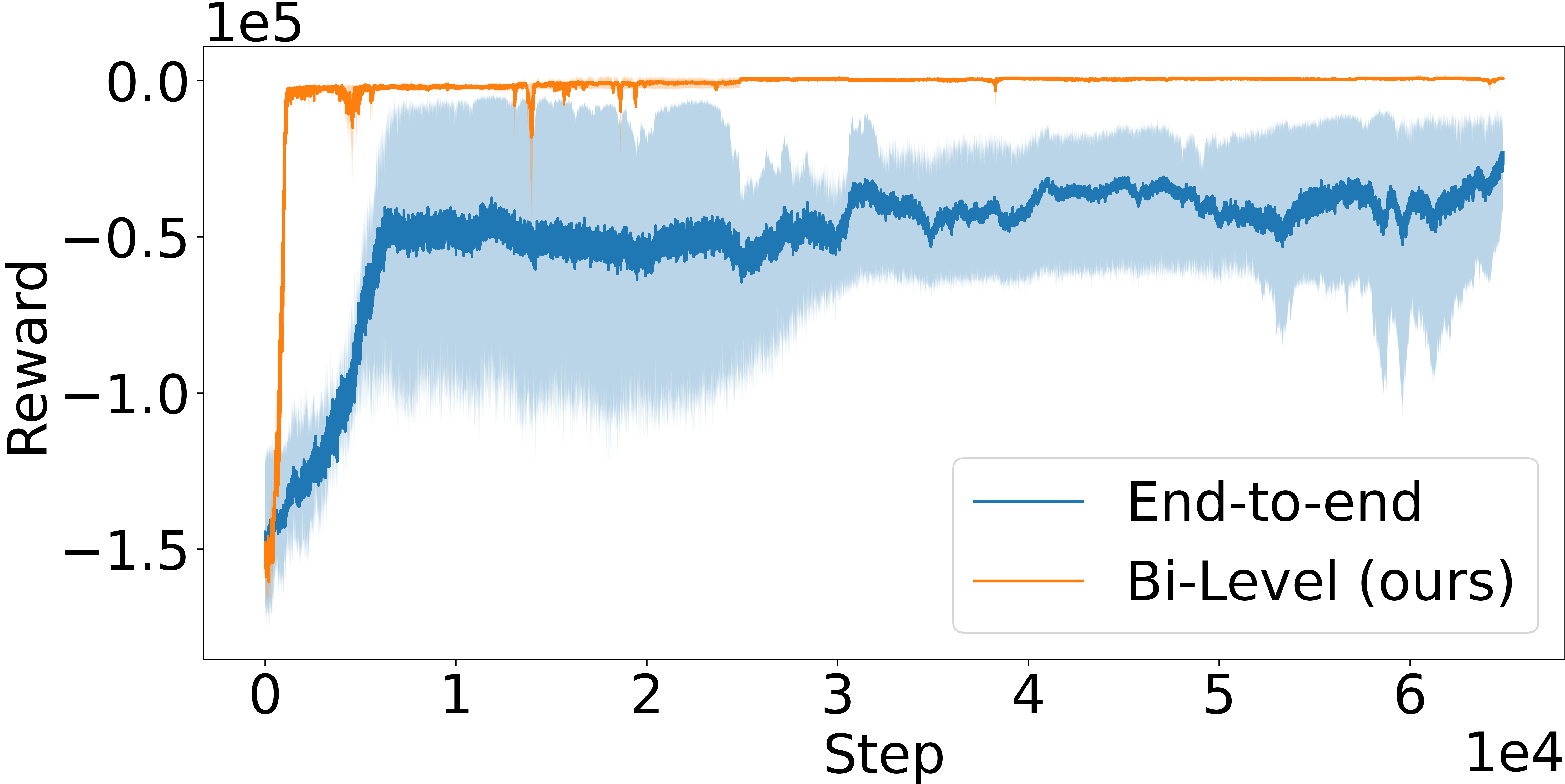}} 
    \subfigure[]{\includegraphics[width=0.33\textwidth]{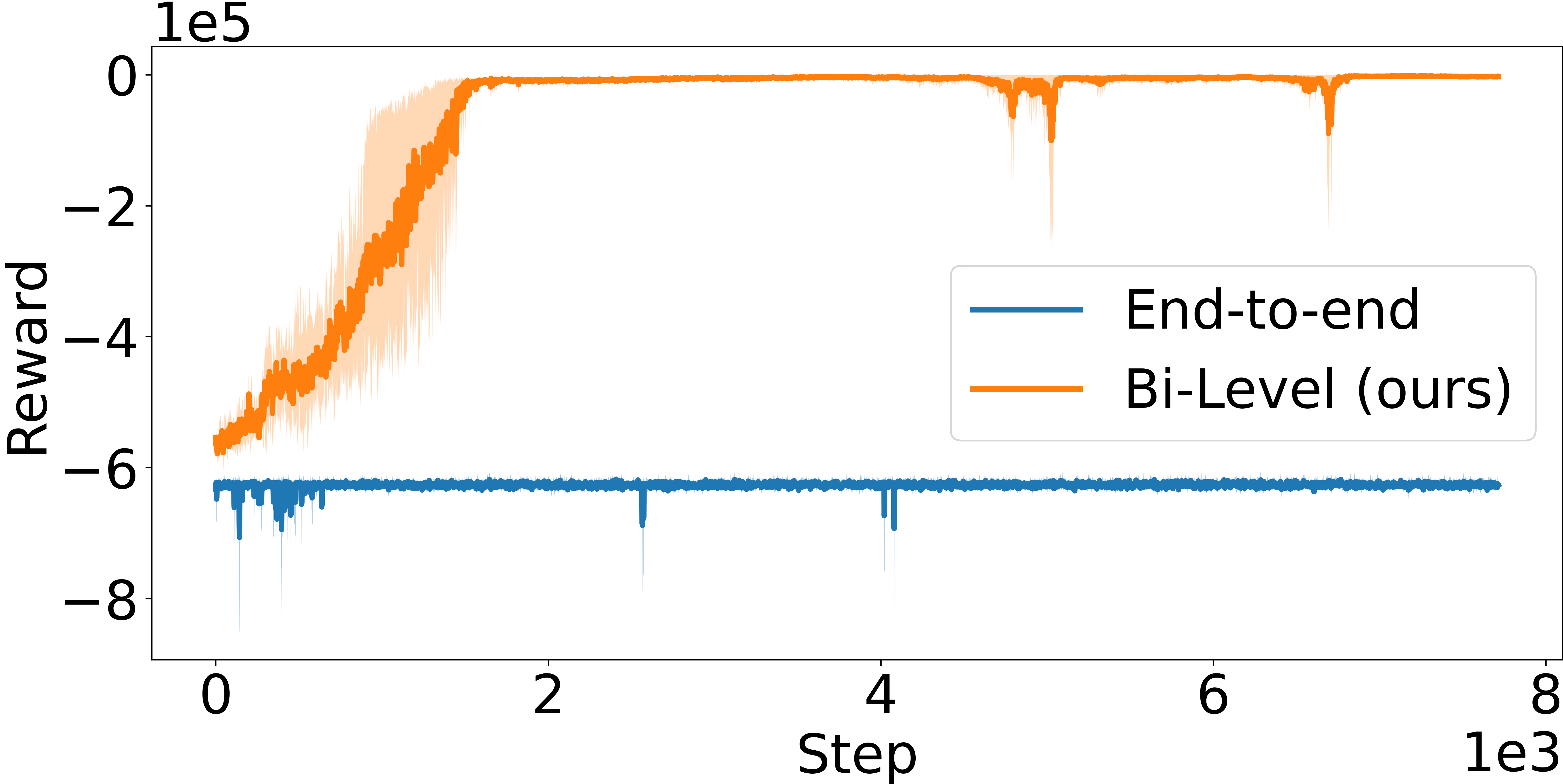}}
    \caption{Learning curve comparison between an RL agent trained end-to-end (blue) and via our bi-level formulation (orange) on the SCIM task (a) 1F2S (b) 1F3S, and (c) 1F10S}
    \label{fig:learning_curve}
\end{figure*}
\textbf{LCP as inductive bias for network computations.} \hspace{2mm}
As a further analysis, we compare with an ablation of our framework, which, as in the majority of literature, is defined as a purely end-to-end RL agent that avoids the LCP and directly maps from environment states to production and shipping actions through either MLPs \cite{PengEtAl2019, OroojlooyjadidEtAl2022} or GNNs. 
Results in Figure \ref{fig:learning_curve} clearly highlight how the bi-level formulation exhibits significantly improved sample efficiency and performance compared to its end-to-end counterpart, which is either substantially slower at converging to good-quality solutions or does not converge at all, as in Figure \ref{fig:learning_curve} (c).
We argue that this behavior is due to two main factors: (1) the bi-level agent operates on a lower-dimensional and well-structured representation via $\hat \st^{t+1}$, and (2) the bi-level formulation provides an implicit inductive bias towards feasible, high-quality solutions via the definition of the LCP. 
Together, these two properties define an RL agent that exhibits improved efficiency and performance. 

\subsection{Dynamic Vehicle Routing}
\begin{table*}[t]
\centering
\footnotesize
\caption{Impact of implicit planning via desired next states.}
\begin{tabular}{l l l c c}
    &  &  & \multicolumn{1}{p{4cm}}{\centering Greedy \\ (i.e., $\argmin_{\ac^t} - \rew(\st^t, \ac^t)$)} & \multicolumn{1}{p{6cm}}{\centering Graph-RL \\ (i.e., $\argmin_{\ac^t} \dist(\hat{\st}^{t+1}, {\st}^{t+1}) - \rew(\st^t, \ac^t)$)} \\
    \midrule 
    \multirow{6}{*}{SCIM} & \multirow{2}{*}{1F2S} & Reward & -102,919 ($\pm$ 2,767) & \textbf{192 ($\pm$ 119)} \\
     &  & \%Oracle & N.A. & \textbf{96.8\%} \\
     & \multirow{2}{*}{1F3S} & Reward & -169,433 ($\pm$ 2,880) & \textbf{997 ($\pm$ 109)} \\
     &  & \%Oracle & N.A. & \textbf{96.0\%} \\
     & \multirow{2}{*}{1F10S} & Reward & -587,661 ($\pm$ 3,862) & \textbf{890 ($\pm$ 288)} \\
     &  & \%Oracle & N.A. & \textbf{99.5\%} \\
    \midrule
    \multirow{6}{*}{DVR} & \multirow{3}{*}{New York} & Reward & 13,978 ($\pm$ 391) & \textbf{15,481} ($\pm$ 397) \\
     &  & Served Demand & 1,357 ($\pm$ 92) & \textbf{1,824} ($\pm$ 87) \\
     &  & \%Oracle & 90.13\% & \textbf{94.9\%} \\
     & \multirow{3}{*}{Shenzhen} & Reward & 35,996 ($\pm$ 499) & \textbf{36,918} ($\pm$ 616) \\
     &  & Served Demand & 2,881 ($\pm$ 98) &  \textbf{3,310} ($\pm$ 92)\\
     &  & \%Oracle & 79.27\% & \textbf{83.9\%} \\
    \bottomrule
    \end{tabular}%
\label{tab:greedy}%
\end{table*}
In the second real-world experiment, we apply our framework to the field of mobility.
Specifically, we focus on the dynamic vehicle routing (DVR) problem, which describes the task of finding the least-cost routes for a fleet of vehicles such that it can satisfy the demand of a set of customers geographically dispersed in a dynamic, stochastic network.
Towards this aim, we consider a transportation network $\graph = \{\nodes, \edges\}$ with $M$ single-occupancy vehicles, where $\nodes$ represents the set of stations (e.g., pickup or drop-off locations) and $\edges$ represents the set of links in the transportation network (e.g., roads), each characterized by a travel time $t_{ij}$ and cost $m_{ij}$. 
At each time step, customers arrive at their origin stations and wait for idle vehicles $\inventory_i$ to transport them to their destinations.
The trip from station $i \in \nodes$ to station $j \in \nodes$ at time $t$ is characterized by a demand $\demand_{ij}^t$ and a price $\price_{ij}$, passengers not served by any vehicle will leave the system and revenue from their trips will be lost. 
The system operator coordinates a fleet of vehicles to best serve the demand for transportation while minimizing the cost of operations.
Concretely, the operator achieves this by controlling the passenger flow $f_{ij,P}^t$ (i.e., vehicles delivering passengers to their destination) and the rebalancing flow $f_{ij,R}^t$ (i.e., vehicles not assigned to passengers and used, for example, to anticipate future demand) at each time step $t$.
Please refer to Appendix \ref{appendix:subsec:dvr} for further details.

\textbf{DVR Markov decision process.} \hspace{2mm}
We formulate the DVR MDP through the following elements:

\textit{Action space ($\acspace$):} we compute the rebalancing flow $\flow_{ij,R}^t$, such that $\ba^t = \{\flow_{ij,R}^t\}_{(i,j) \in \edges}$.
Without loss of generality, we assume the passenger flow is assigned through some independent routine, although the ideas described in this section can be extended to include also passenger flows. 

\textit{Reward $\left(R(\st^t, \ac^t)\right)$:} we select the reward function in the MDP as the operator profit, computed as the difference between trip revenues and operation-related costs. 

\textit{State space ($\stspace$):} the transportation network is described via node features such as the current and projected availability of idle vehicles in each station, current and estimated demand, and provider-level information, e.g., trip price.

\textbf{Bi-Level formulation.} \hspace{2mm}
We further describe an additional instantiation of our bi-level framework for the DVR problem.
First, we define the desired next state $\hat \st^{t+1}$ to represent the desired number of idle vehicles in all stations $\hat \inventory_i^t, \forall i \in \nodes$.
The second step further entails the solution of the LCP to transform the desired  number of idle vehicles into feasible environment actions (i.e., rebalancing flows). 
At a high level, the LCP aims to minimize rebalancing costs while satisfying domain-related constraints  such as ensuring that the total rebalancing flow from a region is upper-bounded by the number of idle vehicles in that region and non-negative.
Please refer to Appendix \ref{appendix:subsubsec:dvr_lcp} for further details.

\textbf{Vehicle routing via network flow.} \hspace{2mm}
We evaluate the algorithms on two real-world urban mobility scenarios based on the cities of New York, USA, and Shenzhen, China.
Results in Table \ref{tab:real_world_experiments_dvr} show how Graph-RL is able to achieve close-to-optimal performance in both environments.
Specifically, the vehicle routing policies learned through Graph-RL achieve 94.9\% (New York) and 83.8\% (Shenzhen) of oracle performance, while showing a 23.3\% (New York) and 33.7\% (Shenzhen) increase in operator profit compared to the domain-driven heuristic, which attempts to preserve equal access to vehicles across stations in the transportation network.
As observed for SCIM problems, the results confirm that end-to-end RL approaches struggle with high-dimensional action spaces ($75$ and $90$ edges in New York and Shenzhen environments, respectively) and fail to learn effective routing strategies.
Lastly, to assess the transferability and generalization capabilities of Graph-RL, we study the extent to which policies can be trained on one city and later applied to the other without further training (i.e., zero-shot). 
Table \ref{tab:real_world_experiments_dvr} shows that routing policies learned in one city exhibit a promising degree of portability to novel environments, with only minimal performance decay.
As introduced in Section \ref{subsec:importance_of_gnns}, this experiment further highlights the importance of the locality of graph network-based policies: by learning a shared, local operator, policies learned through graph-RL can potentially be applied to arbitrary graph topologies.
Crucially, policies with structural transfer capabilities could enable system operators to re-use previous experience, thus avoiding expensive re-training when exposed to new problem instances. 

\subsection{Comparison to Greedy Planning}
\label{subsec:distance_metric}
The role of the distance metric (and the generated desired next state) in Eq. \eqref{eq:reb_obj} is to capture the value of future reward in the greedy one-step inner optimization problem, ultimately allowing for implicit long-term planning (please refer to Appendix \ref{appendix:subsec:distance_metric} for a broader discussion).
To quantify this intuition, in Table \ref{tab:greedy} we compare the proposed bi-level approach to a \textit{greedy} policy that acts optimally with respect to the one-step optimization problem. 
Concretely, if on one hand the proposed bi-level approach attempts to achieve as best as possible the desired next state (i.e., $\argmin_{\ac^t} \dist(\hat{\st}^{t+1}, {\st}^{t+1}) - \rew(\st^t, \ac^t)$), the greedy policy ignores the distance term and optimizes solely short-term reward (i.e., $\argmin_{\ac^t}  - \rew(\st^t, \ac^t)$)).
Results in Table \ref{tab:greedy} highlight how the presence of the desired next state, and ultimately, of the bi-level approach, is instrumental in achieving effective long-term performance. 
Crucially, since both producing a commodity (SCIM) and rebalancing a vehicle (DVR) are only defined by negative rewards, these only indirectly participate to long-term positive reward via a better (i) product availability or (ii) positioning of vehicles, and thus cannot be measured by the one-step optimization problem.
This results in the greedy policy (i) being unable to fulfill any demand in the SCIM problem and (ii) achieving lower profit in the DVR problem. 
It is important to highlight how, in the DVR problem, the greedy policy achieves reasonably good reward (i.e., profit) because the system can partially self-sustain itself only through passenger trips. 
However, greediness causes the number of served customers to be considerably smaller, with Graph-RL achieving $\approx$ +35\% in New York and $\approx$ +15\% in Shenzhen, thus clearly showing the benefit of optimizing for long-term reward via the minimization of the distance metric.

\section{Conclusion}
\label{sec:conclusion}
Research in network optimization problems, in both theory and practice, is largely scattered across the control, management science, and optimization literature, potentially hindering scientific progress.
In this work, we propose a general framework that could enable learning-based approaches to help address the open challenges in this space: handling nonlinear dynamics and scalability, among others.
Specifically, instead of approaching the problem through pure end-to-end reinforcement learning, we introduced a general bi-level formulation that leverages the specific strengths of direct optimization, reinforcement learning, and graph representation learning. 
Our approach shows strong performance on all problem settings we evaluate, substantially outperforming both optimization-based and RL-based approaches.
In future work, we plan to investigate ways to exploit the non-parametric nature of our approach and take a step in the direction of learning generalist graph optimizers. 
More generally, we believe this research opens several promising directions for the extension of these concepts to a broader class of large-scale, real-world applications.



\bibliography{ASL_papers, main}

\newcommand{\noopsort}[1]{} \newcommand{\printfirst}[2]{#1}
  \newcommand{\singleletter}[1]{#1} \newcommand{\switchargs}[2]{#2#1}
\begin{thebibliography}{68}
\providecommand{\natexlab}[1]{#1}
\providecommand{\url}[1]{\texttt{#1}}
\expandafter\ifx\csname urlstyle\endcsname\relax
  \providecommand{\doi}[1]{doi: #1}\else
  \providecommand{\doi}{doi: \begingroup \urlstyle{rm}\Url}\fi

\bibitem[Agrawal et~al.(2019{\natexlab{a}})Agrawal, Barratt, Boyd, Busseti, and
  Moursi]{AgrawalBarrattEtAl2019}
Agrawal, A., Barratt, S., Boyd, S., Busseti, E., and Moursi, W.~M.
\newblock Differentiating through a conic program.
\newblock \emph{{Journal of Applied and Numerical Optimization}}, 1\penalty0
  (2):\penalty0 107--115, 2019{\natexlab{a}}.

\bibitem[Agrawal et~al.(2019{\natexlab{b}})Agrawal, Barratt, Boyd, and
  Stellato]{AgrawalBarrattEtAl2019b}
Agrawal, A., Barratt, S., Boyd, S., and Stellato, B.
\newblock Learning convex optimization control policies.
\newblock In \emph{{Learning for Dynamics \& Control}}, 2019{\natexlab{b}}.

\bibitem[Amos \& Kolter(2017)Amos and Kolter]{AmosKolter2017}
Amos, B. and Kolter, J.~Z.
\newblock {OptNet}: Differentiable optimization as a layer in neural networks.
\newblock In \emph{{Int.\ Conf.\ on Machine Learning}}, 2017.

\bibitem[Amos \& Yarats(2020)Amos and Yarats]{amos2020differentiable}
Amos, B. and Yarats, D.
\newblock The differentiable cross-entropy method.
\newblock In \emph{{Int.\ Conf.\ on Machine Learning}}, pp.\  291--302, 2020.

\bibitem[Amos et~al.(2018)Amos, Jimenez, Sacks, Boots, and
  Kolter]{amos2018differentiable}
Amos, B., Jimenez, I., Sacks, J., Boots, B., and Kolter, J.~Z.
\newblock Differentiable mpc for end-to-end planning and control.
\newblock \emph{{Conf.\ on Neural Information Processing Systems}}, 31, 2018.

\bibitem[Astrom(2012)]{aastrom2012introduction}
Astrom, K.~J.
\newblock \emph{Introduction to stochastic control theory}.
\newblock Courier Corporation, 2012.

\bibitem[Bellamy \& Basole(2013)Bellamy and Basole]{bellamy2013network}
Bellamy, M.~A. and Basole, R.~C.
\newblock Network analysis of supply chain systems: A systematic review and
  future research.
\newblock \emph{Systems Engineering}, 16\penalty0 (2):\penalty0 235--249, 2013.

\bibitem[Bertsekas(1995)]{Bertsekas1995}
Bertsekas, D.
\newblock \emph{Dynamic programming and optimal control}.
\newblock {Athena Scientific}, first edition, 1995.

\bibitem[Bertsekas(2019)]{bertsekas2019reinforcement}
Bertsekas, D.
\newblock \emph{Reinforcement learning and optimal control}.
\newblock Athena Scientific, 2019.

\bibitem[Bertsekas \& Tsitsiklis(1996)Bertsekas and
  Tsitsiklis]{bertsekas1996neuro}
Bertsekas, D. and Tsitsiklis, J.~N.
\newblock \emph{Neuro-dynamic programming}.
\newblock Athena Scientific, 1996.

\bibitem[Bienstock et~al.(2014)Bienstock, Chertkov, and
  Harnett]{bienstock2014chance}
Bienstock, D., Chertkov, M., and Harnett, S.
\newblock Chance-constrained optimal power flow: Risk-aware network control
  under uncertainty.
\newblock \emph{{SIAM} Review}, 56\penalty0 (3):\penalty0 461--495, 2014.

\bibitem[Birge \& Louveaux(2011)Birge and Louveaux]{BirgeLouveaux2011}
Birge, J.~R. and Louveaux, F.
\newblock \emph{Introduction to stochastic programming}.
\newblock {Springer Science \& Business Media}, 2011.

\bibitem[Boyd \& Vandenberghe(2004)Boyd and Vandenberghe]{BoydVandenberghe2004}
Boyd, S. and Vandenberghe, L.
\newblock \emph{Convex Optimization}.
\newblock {Cambridge Univ.\ Press}, 2004.

\bibitem[Daganzo(1978)]{Daganzo1978}
Daganzo, C.-F.
\newblock An approximate analytic model of many-to-many demand responsive
  transportation systems.
\newblock \emph{{Transportation Research}}, 12\penalty0 (5):\penalty0 325--333,
  1978.

\bibitem[Dantzig(1982)]{dantzig1982reminiscences}
Dantzig, G.~B.
\newblock Reminiscences about the origins of linear programming.
\newblock \emph{Operations Research Letters}, 1\penalty0 (2):\penalty0 43--48,
  1982.

\bibitem[Dinh \& Diehl(2010)Dinh and Diehl]{DinhDiehl2010}
Dinh, Q.~T. and Diehl, M.
\newblock Local convergence of sequential convex programming for nonconvex
  optimization.
\newblock In \emph{Recent Advances in Optimization and its Applications in
  Engineering}. {Springer}, 2010.

\bibitem[Dommel \& Tinney(1968)Dommel and Tinney]{dommel1968optimal}
Dommel, H.~W. and Tinney, W.~F.
\newblock Optimal power flow solutions.
\newblock \emph{IEEE Transactions on power apparatus and systems}, \penalty0
  (10):\penalty0 1866--1876, 1968.

\bibitem[Donti et~al.(2017)Donti, Amos, and Kolter]{donti2017task}
Donti, P., Amos, B., and Kolter, J.~Z.
\newblock Task-based end-to-end model learning in stochastic optimization.
\newblock \emph{{Conf.\ on Neural Information Processing Systems}}, 30, 2017.

\bibitem[Dumouchelle et~al.(2022)Dumouchelle, Patel, Khalil, and
  Bodur]{dumouchelle2022neur2sp}
Dumouchelle, J., Patel, R., Khalil, E.~B., and Bodur, M.
\newblock Neur2sp: Neural two-stage stochastic programming.
\newblock \emph{arXiv:2205.12006}, 2022.

\bibitem[Finn et~al.(2017)Finn, Abbeel, and Levine]{FinnAbbeelEtAl2017}
Finn, C., Abbeel, P., and Levine, S.
\newblock Model-agnostic meta-learning for fast adaptation of deep networks.
\newblock In \emph{{Int.\ Conf.\ on Machine Learning}}, 2017.

\bibitem[Flood(1997)]{flood1997telecommunication}
Flood, J.~E.
\newblock \emph{Telecommunication networks}.
\newblock IET, 1997.

\bibitem[Ford \& Fulkerson(1956)Ford and Fulkerson]{ford1956maximal}
Ford, L.~R. and Fulkerson, D.~R.
\newblock Maximal flow through a network.
\newblock \emph{Canadian journal of Mathematics}, 8:\penalty0 399--404, 1956.

\bibitem[Ford \& Fulkerson(1958)Ford and Fulkerson]{FordFulkerson1958}
Ford, L.~R. and Fulkerson, D.~R.
\newblock Constructing maximal dynamic flows from static flows.
\newblock \emph{Operations Research}, 6\penalty0 (3):\penalty0 419--433, 1958.

\bibitem[Ford \& Fulkerson(1962)Ford and Fulkerson]{FordFulkerson1962}
Ford, L.~R. and Fulkerson, D.~R.
\newblock \emph{Flows in Networks}.
\newblock {Princeton Univ.\ Press}, 1962.

\bibitem[Fujimoto et~al.(2022)Fujimoto, Meger, Precup, Nachum, and
  Gu]{fujimoto2022should}
Fujimoto, S., Meger, D., Precup, D., Nachum, O., and Gu, S.~S.
\newblock Why should i trust you, bellman? the bellman error is a poor
  replacement for value error.
\newblock \emph{arXiv:2201.12417}, 2022.

\bibitem[Gammelli et~al.(2021)Gammelli, Yang, Harrison, Rodrigues, Pereira, and
  Pavone]{GammelliYangEtAl2021}
Gammelli, D., Yang, K., Harrison, J., Rodrigues, F., Pereira, F.~C., and
  Pavone, M.
\newblock Graph neural network reinforcement learning for autonomous
  mobility-on-demand systems.
\newblock In \emph{{Proc.\ IEEE Conf.\ on Decision and Control}}, 2021.

\bibitem[Gammelli et~al.(2022)Gammelli, Yang, Harrison, Rodrigues, Pereira, and
  Pavone]{GammelliYangEtAl2022}
Gammelli, D., Yang, K., Harrison, J., Rodrigues, F., Pereira, F., and Pavone,
  M.
\newblock Graph meta-reinforcement learning for transferable autonomous
  mobility-on-demand.
\newblock In \emph{{ACM Int.\ Conf.\ on Knowledge Discovery and Data Mining}},
  2022.

\bibitem[Gilmer et~al.(2017)Gilmer, Schoenholz, Riley, Vinyals, and
  Dahl]{GilmerEtAl2017}
Gilmer, J., Schoenholz, S., Riley, P., Vinyals, O., and Dahl, G.
\newblock Neural message passing for quantum chemistry.
\newblock In \emph{{Int.\ Conf.\ on Machine Learning}}, 2017.

\bibitem[Glynn(1990)]{glynn1990likelihood}
Glynn, P.~W.
\newblock Likelihood ratio gradient estimation for stochastic systems.
\newblock \emph{Communications of the ACM}, 33\penalty0 (10):\penalty0 75--84,
  1990.

\bibitem[Harrison et~al.(2018)Harrison, Sharma, and Pavone]{harrison2018meta}
Harrison, J., Sharma, A., and Pavone, M.
\newblock Meta-learning priors for efficient online bayesian regression.
\newblock In \emph{{Workshop on Algorithmic Foundations of Robotics}}, pp.\
  318--337, 2018.

\bibitem[Hillier \& Lieberman(1995)Hillier and
  Lieberman]{hillier1967introduction}
Hillier, F. and Lieberman, G.
\newblock \emph{Introduction to operations research}.
\newblock 1995.

\bibitem[Huneault \& Galiana(1991)Huneault and Galiana]{huneault1991survey}
Huneault, M. and Galiana, F.~D.
\newblock A survey of the optimal power flow literature.
\newblock \emph{IEEE transactions on Power Systems}, 6\penalty0 (2):\penalty0
  762--770, 1991.

\bibitem[IBM(1987)]{ios_ILOG:1987}
IBM.
\newblock \emph{{ILOG CPLEX} User's guide}.
\newblock IBM ILOG, 1987.

\bibitem[Ichter et~al.(2018)Ichter, Harrison, and Pavone]{ichter2018learning}
Ichter, B., Harrison, J., and Pavone, M.
\newblock Learning sampling distributions for robot motion planning.
\newblock In \emph{{Proc.\ IEEE Conf.\ on Robotics and Automation}}, pp.\
  7087--7094, 2018.

\bibitem[Jakobson \& Weissman(1995)Jakobson and Weissman]{jakobson1995real}
Jakobson, G. and Weissman, M.
\newblock Real-time telecommunication network management: Extending event
  correlation with temporal constraints.
\newblock In \emph{International Symposium on Integrated Network Management},
  pp.\  290--301, 1995.

\bibitem[Key \& Cope(1990)Key and Cope]{key1990distributed}
Key, P.~B. and Cope, G.~A.
\newblock Distributed dynamic routing schemes.
\newblock \emph{IEEE Communications Magazine}, 28\penalty0 (10):\penalty0
  54--58, 1990.

\bibitem[Kipf \& Welling(2017)Kipf and Welling]{KipfWelling2017}
Kipf, T.-N. and Welling, M.
\newblock Semi-supervised classification with graph convolutional networks.
\newblock In \emph{{Int.\ Conf.\ on Learning Representations}}, 2017.

\bibitem[Konda \& Tsitsiklis(1999)Konda and Tsitsiklis]{KondaEtAl1999}
Konda, V. and Tsitsiklis, J.
\newblock Actor-critic algorithms.
\newblock In \emph{{Conf.\ on Neural Information Processing Systems}}, 1999.

\bibitem[Landry et~al.(2019)Landry, Lorenzetti, Manchester, and
  Pavone]{landry2019bilevel}
Landry, B., Lorenzetti, J., Manchester, Z., and Pavone, M.
\newblock Bilevel optimization for planning through contact: A semidirect
  method.
\newblock In \emph{The International Symposium of Robotics Research}, pp.\
  789--804, 2019.

\bibitem[Levine et~al.(2020)Levine, Kumar, Tucker, and Fu]{LevineEtAl2020}
Levine, S., Kumar, A., Tucker, G., and Fu, J.
\newblock Offline reinforcement learning: Tutorial, review, and perspectives on
  open problems.
\newblock \emph{arXiv:2005.01643}, 2020.

\bibitem[Lew et~al.(2022)Lew, Singh, Prats, Bingham, Weisz, Holson, Zhang,
  Sindhwani, Lu, Xia, et~al.]{LewSinghEtAl2022}
Lew, T., Singh, S., Prats, M., Bingham, J., Weisz, J., Holson, B., Zhang, X.,
  Sindhwani, V., Lu, Y., Xia, F., et~al.
\newblock Robotic table wiping via reinforcement learning and whole-body
  trajectory optimization.
\newblock \emph{arXiv preprint arXiv:2210.10865}, 2022.

\bibitem[Li \& Bo(2007)Li and Bo]{li2007dcopf}
Li, F. and Bo, R.
\newblock Dcopf-based lmp simulation: algorithm, comparison with acopf, and
  sensitivity.
\newblock \emph{IEEE Transactions on Power Systems}, 22\penalty0 (4):\penalty0
  1475--1485, 2007.

\bibitem[Metz et~al.(2019)Metz, Maheswaranathan, Nixon, Freeman, and
  Sohl-Dickstein]{metz2019understanding}
Metz, L., Maheswaranathan, N., Nixon, J., Freeman, D., and Sohl-Dickstein, J.
\newblock Understanding and correcting pathologies in the training of learned
  optimizers.
\newblock In \emph{{Int.\ Conf.\ on Machine Learning}}, pp.\  4556--4565, 2019.

\bibitem[Metz et~al.(2022)Metz, Harrison, Freeman, Merchant, Beyer, Bradbury,
  Agrawal, Poole, Mordatch, Roberts, et~al.]{metz2022velo}
Metz, L., Harrison, J., Freeman, C.~D., Merchant, A., Beyer, L., Bradbury, J.,
  Agrawal, N., Poole, B., Mordatch, I., Roberts, A., et~al.
\newblock Velo: Training versatile learned optimizers by scaling up.
\newblock \emph{arXiv preprint arXiv:2211.09760}, 2022.

\bibitem[Mnih et~al.(2015)Mnih, Kavukcuoglu, Silver, Rusu,
  et~al.]{MnihKavukcuogluEtAl2015}
Mnih, V., Kavukcuoglu, K., Silver, D., Rusu, A.~A., et~al.
\newblock Human-level control through deep reinforcement learning.
\newblock \emph{{Nature}}, 518\penalty0 (7540):\penalty0 529--533, 2015.

\bibitem[Mnih et~al.(2016)Mnih, Puigdomenech, Mirza, Graves, Lillicrap, Harley,
  Silver, and Kavukcuoglu]{MnihPuigdomenechEtAl2016}
Mnih, V., Puigdomenech, A., Mirza, M., Graves, A., Lillicrap, T.-P., Harley,
  T., Silver, D., and Kavukcuoglu, K.
\newblock Asynchronous methods for deep reinforcement learning.
\newblock In \emph{{Int.\ Conf.\ on Learning Representations}}, 2016.

\bibitem[Murota(2009)]{Murota2009}
Murota, K.
\newblock \emph{Matrices and Matroids for Systems Analysis}.
\newblock {Springer Science \& Business Media}, 1 edition, 2009.

\bibitem[Oroojlooyjadid et~al.(2022)Oroojlooyjadid, Nazari, Snyder, and
  Tak{\'a}{\v{c}}]{OroojlooyjadidEtAl2022}
Oroojlooyjadid, A., Nazari, M., Snyder, L.~V., and Tak{\'a}{\v{c}}, M.
\newblock A deep q-network for the beer game: Deep reinforcement learning for
  inventory optimization.
\newblock \emph{Manufacturing and Service Operations Management}, 24\penalty0
  (1):\penalty0 285--304, 2022.

\bibitem[Paszke et~al.(2019)Paszke, Gross, Massa, Lerer,
  et~al.]{PaszkeGrossEtAl2019}
Paszke, A., Gross, S., Massa, F., Lerer, A., et~al.
\newblock Pytorch: An imperative style, high-performance deep learning library.
\newblock \emph{arXiv preprint arXiv:1912.01703}, 2019.

\bibitem[Peng et~al.(2019)Peng, Zhang, Feng, Zhang, Wu, and Su]{PengEtAl2019}
Peng, Z., Zhang, Y., Feng, Y., Zhang, T., Wu, Z., and Su, H.
\newblock Deep reinforcement learning approach for capacitated supply chain
  optimization under demand uncertainty.
\newblock In \emph{2019 Chinese Automation Congress (CAC)}, 2019.

\bibitem[Pereira \& Pinto(1991)Pereira and Pinto]{pereira1991multi}
Pereira, M.~V. and Pinto, L.~M.
\newblock Multi-stage stochastic optimization applied to energy planning.
\newblock \emph{{Mathematical Programming}}, 52\penalty0 (1):\penalty0
  359--375, 1991.

\bibitem[Popovskij et~al.(2011)Popovskij, Barkalov, and
  Titarenko]{popovskij2011control}
Popovskij, V., Barkalov, A., and Titarenko, L.
\newblock \emph{Control and adaptation in telecommunication systems:
  Mathematical Foundations}, volume~94.
\newblock Springer Science \& Business Media, 2011.

\bibitem[Powell(2022)]{powell2022reinforcement}
Powell, W.~B.
\newblock \emph{Reinforcement Learning and Stochastic Optimization: A unified
  framework for sequential decisions}.
\newblock Wiley, 2022.

\bibitem[Power \& Berenson(2022)Power and Berenson]{power2022variational}
Power, T. and Berenson, D.
\newblock Variational inference mpc using normalizing flows and
  out-of-distribution projection.
\newblock \emph{arXiv:2205.04667}, 2022.

\bibitem[Rawlings \& Mayne(2013)Rawlings and Mayne]{RawlingsMayne2013}
Rawlings, J. and Mayne, D.
\newblock \emph{Model predictive control: Theory and design}.
\newblock {Nob Hill Publishing}, 2013.

\bibitem[Sacks \& Boots(2022)Sacks and Boots]{sacks2022learning}
Sacks, J. and Boots, B.
\newblock Learning to optimize in model predictive control.
\newblock In \emph{{Proc.\ IEEE Conf.\ on Robotics and Automation}}, pp.\
  10549--10556, 2022.

\bibitem[Sarimveis et~al.(2008)Sarimveis, Patrinos, Tarantilis, and
  Kiranoudis]{sarimveis2008dynamic}
Sarimveis, H., Patrinos, P., Tarantilis, C.~D., and Kiranoudis, C.~T.
\newblock Dynamic modeling and control of supply chain systems: A review.
\newblock \emph{Computers \& operations research}, 35\penalty0 (11):\penalty0
  3530--3561, 2008.

\bibitem[Shapiro et~al.(2014)Shapiro, Dentcheva, and
  Ruszczy{\'n}ski]{ShapiroDentchevaEtAl2014}
Shapiro, A., Dentcheva, D., and Ruszczy{\'n}ski, A.
\newblock \emph{Lectures on stochastic programming: Modeling and theory}.
\newblock {SIAM}, second edition, 2014.

\bibitem[Sutton \& Barto(1998)Sutton and Barto]{SuttonBarto1998}
Sutton, R.~S. and Barto, A.~G.
\newblock \emph{Reinforcement Learning: An Introduction}.
\newblock {MIT Press}, 1 edition, 1998.

\bibitem[Tamar et~al.(2017)Tamar, Thomas, Zhang, Levine, and
  Abbeel]{tamar2017learning}
Tamar, A., Thomas, G., Zhang, T., Levine, S., and Abbeel, P.
\newblock Learning from the hindsight plan—episodic mpc improvement.
\newblock In \emph{{Proc.\ IEEE Conf.\ on Robotics and Automation}}, pp.\
  336--343, 2017.

\bibitem[Van~de Wiele et~al.(2020)Van~de Wiele, Warde-Farley, Mnih, and
  Mnih]{van2020q}
Van~de Wiele, T., Warde-Farley, D., Mnih, A., and Mnih, V.
\newblock Q-learning in enormous action spaces via amortized approximate
  maximization.
\newblock \emph{arXiv:2001.08116}, 2020.

\bibitem[Van~Roy et~al.(1997)Van~Roy, Bertsekas, Lee, and
  Tsitsiklis]{VanRoyEtAl1997}
Van~Roy, B., Bertsekas, D.~P., Lee, Y., and Tsitsiklis, J.~N.
\newblock A neuro-dynamic programming approach to retailer inventory
  management.
\newblock In \emph{{Proc.\ IEEE Conf.\ on Decision and Control}}, 1997.

\bibitem[Veli{\v c}kovi{\'c} et~al.(2018)Veli{\v c}kovi{\'c}, Cucurull,
  Casanova, Romero, Li{\`o}, and Bengio]{Velickovic2018}
Veli{\v c}kovi{\'c}, P., Cucurull, G., Casanova, A., Romero, A., Li{\`o}, O.,
  and Bengio, Y.
\newblock Graph attention networks.
\newblock In \emph{{Int.\ Conf.\ on Learning Representations}}, 2018.

\bibitem[Wang et~al.(2018)Wang, Szeto, Han, and Friesz]{WangSzetoEtAl2018}
Wang, Y., Szeto, W.~Y., Han, K., and Friesz, T.
\newblock Dynamic traffic assignment: A review of the methodological advances
  for environmentally sustainable road transportation applications.
\newblock \emph{{Transportation Research Part B: Methodological}},
  111:\penalty0 370--394, 2018.

\bibitem[Williams(1992)]{Williams1992}
Williams, R.-J.
\newblock Simple statistical gradient-following algorithms for connectionist
  reinforcement learning.
\newblock \emph{{Machine Learning}}, 1992.

\bibitem[Xiao et~al.(2022)Xiao, Zhang, Choromanski, Lee, Francis, Varley, Tu,
  Singh, Xu, Xia, Takayama, Frostig, Tan, Parada, and
  Sindhwani]{xiao2022learning}
Xiao, X., Zhang, T., Choromanski, K.~M., Lee, T.-W.~E., Francis, A., Varley,
  J., Tu, S., Singh, S., Xu, P., Xia, F., Takayama, L., Frostig, R., Tan, J.,
  Parada, C., and Sindhwani, V.
\newblock Learning model predictive controllers with real-time attention for
  real-world navigation.
\newblock In \emph{{Conf.\ on Robot Learning}}, 2022.

\bibitem[Xu et~al.(2020)Xu, Li, Zhang, Du, Kawarabayashi, and
  Jegelka]{XuEtAl2020}
Xu, K., Li, J., Zhang, M., Du, S., Kawarabayashi, K., and Jegelka, S.
\newblock What can neural networks reason about?
\newblock In \emph{{Int.\ Conf.\ on Learning Representations}}, 2020.

\bibitem[Zhang et~al.(2016)Zhang, Rossi, and Pavone]{ZhangRossiEtAl2016b}
Zhang, R., Rossi, F., and Pavone, M.
\newblock Model predictive control of {Autonomous} {Mobility-on-Demand}
  systems.
\newblock In \emph{{Proc.\ IEEE Conf.\ on Robotics and Automation}}, 2016.

\end{thebibliography}
\bibliographystyle{icml2023}

\newpage
\appendix
\onecolumn
\section{Dynamic Network Control}
\label{appendix:dynamic_network_control}

In this section, we make concrete our discussion on nonlinear problem formulations for network control problems.

\paragraph{Elements violating the linearity assumption}
Real-world systems are characterized by many factors that cannot be reliably modeled through the linear problem described in Section \ref{sec:problem}.
In what follows, we discuss a (non-exhaustive) list of factors potentially breaking such linearity assumptions:
\begin{itemize}
    \item \textbf{Stochasticity.} Various stochastic elements can impact the problem. Commodity transitions in Section \ref{subsec:the_linear_control_problem} were defined as being deterministic; in practice in many problems, there are elements of stochasticity to these transitions. For example, random demand may reduce supply by an unpredictable amount; vehicles may be randomly added in a transportation problem; and packages may be lost in a supply chain setting. In addition to these state transitions, constraints may be stochastic as well: flow times or edge capacities may be stochastic, as when a road is shared with other users, or costs for flows and exchange may be stochastic. 
    \item \textbf{Nonlinearity.} Various elements of the state evolution, constraints, or cost function may be nonlinear. The objective may be chosen to be a risk-sensitive or robust metric applied to the distribution of outcomes, as is common in financial problems. The state evolution may have natural saturating behavior (e.g. automatic load shedding). Indeed, many real constraints will have natural nonlinear behavior.
    \item \textbf{Time-varying costs and constraints.} Similar to the stochastic case, various quantities may be time-varying. However, it is possible that they are time-varying in a structured way, as opposed to randomly. For example, demand for transportation may vary over the time of day, or purchasing costs may vary over the year. 
    \item \textbf{Unknown dynamics elements.} While not a major focus of discussion in the paper up to this point, elements of the underlying dynamics may be partially or wholly unknown. Our reinforcement learning formulation is capapble of addressing this by learning policies directly from data, in contrast to standard control techniques. 
\end{itemize}

\section{Methodology}
\label{appendix:methodology}
In this section, we discuss network architectures and RL components more in detail. 

\subsection{Network Architecture}
\label{appendix:subsec:network_architeture}
Specifically, we first introduce the basic building blocks of our graph neural network architecture.
Let us define with $\mathbf{x}_i \in \mathbb{R}^{D_{\mathbf{x}}}$ and $\mathbf{e}_{ji} \in \mathbb{R}^{D_{\mathbf{e}}}$ the $D_{\mathbf{x}}$-dimensional vector of node features of node $i$ and the $D_{\mathbf{e}}$-dimensional vector of edge features from node $j$ to node $i$, respectively.

We define the update function of node features through either:

\begin{itemize}
    \item Message passing neural network (MPNN) \cite{GilmerEtAl2017}  defined as
    \begin{equation}
       \mathbf{x}_i^{(k)} = \bigoplus_{j \in \mathcal{N}^{-}(i)} f_{\theta}\left(\mathbf{x}_i^{(k-1)}, \mathbf{x}_j^{(k-1)}, \mathbf{e}_{ji} \right),
       \label{eq:mpnn}
    \end{equation}
    where $k$ indicates the $k$-th layer of message passing in the GNN with $k=0$ indicating raw environment features, i.e., $\mathbf{x}^{(0)}_i = \mathbf{x}_i$, and $\bigoplus$ denotes a differentiable, permutation invariant function, e.g., sum, mean or max.

    \item Graph convolution network (GCN) \cite{KipfWelling2017}  defined as
    \begin{equation}
        \mathbf{X'} = f(\mathbf{X}, \bA) = \sigma\left(\hat{\bD}^{-\frac{1}{2}} \hat{\bA} \hat{\bD}^{-\frac{1}{2}} \mathbf{X} \bW\right), 
        \label{eq:gcn}
    \end{equation}
    where $\mathbf{X}$ is the $N_v \times D_{\mathbf{x}}$ feature matrix, $\bA$ is the adjacency matrix with $\hat{\bA} = \bA + \bI$ and $\bI$ is the identity matrix. $\hat{\bD}$ is the diagonal node degree matrix of $\hat{\bA}$, $\sigma(\cdot)$ is a non-linear activation function (e.g., ReLU) and  $\bW$ is a matrix of learnable parameters.
\end{itemize}
We select the specific architecture based on the alignment with the problem characteristics.
We note that these network architectures can be used to define both policy and value function estimator, depending on the reinforcement learning algorithm of interest (e.g., actor-critic \cite{KondaEtAl1999}, value-based \cite{MnihKavukcuogluEtAl2015}, etc.).
As an example, in our implementation, we define two separate decoder architectures for the actor and critic networks of an Advantage Actor Critic (A2C) \cite{MnihPuigdomenechEtAl2016} algorithm.
Below is a summary of the specific architectures used in this work:
\begin{itemize}
    \item Section \ref{subsec:minimum_cost_flow}. We use an MPNN as in \eqref{eq:mpnn}, with a $\max$ aggregation function i.e., $\bigoplus = \max$.
    We define the output of our policy network to represent the concentration parameters $\mathbf{\alpha} \in \mathbb{R}_{+}^{N_v}$ of a Dirichlet distribution, such that $\ba_t \sim \text{Dir}(\ba_t | \mathbf{\alpha})$, and where the positivity of $\mathbf{\alpha}$ is ensured by a Softplus nonlinearity.
    On the other hand, the critic is characterized by a \emph{global} sum-pooling performed after $K$ layers of MPNN.

    \item Section \ref{subsec:supply_chain_inventory_management}. We use an MPNN as in \eqref{eq:mpnn}, with a $\text{sum}$ aggregation function i.e., $\bigoplus = \text{sum}$.
    We define the output of our policy network to represent the (1) concentration parameters $\mathbf{\alpha} \in \mathbb{R}_{+}^{|\storenodes|}$ of a Dirichlet distribution for computing the flow actions, and (2) mean $\mathbf{\mu} \in \mathbb{R}^{|\warehousenodes|}$ and standard deviation $\mathbf{\sigma} \in \mathbb{R}_{+}^{|\warehousenodes|}$ of a Gaussian distribution for the production action.
    On the other hand, the critic is characterized by a \emph{global} sum-pooling performed after $K$ layers of MPNN.

    \item Section \ref{subsec:dynamic_vehicle_routing}. We use a GCN as in \eqref{eq:gcn}. Actor and critic outputs are defined as in the minimum cost flow problem. 
\end{itemize}

\paragraph{Handling dynamic topologies.} A defining property of our framework is its ability to deal with time-dependent graph connectivity (e.g., edges or nodes are added/dropped during the course of an episode).
Specifically, our framework achieves this by (i) considering the problem as a one-step decision-making problem, i.e., avoiding the dependency on potentially unknown future topologies, and (ii) exploiting the capacity of GNNs to handle diverse graph topologies.
Crucially, no matter the current state of the graph, GNN-based agents are capable of computing a desired next state for the network, which will then be converted into actionable flow decisions by the LCP.

\subsection{RL Details}
We further discuss practical aspects within our bi-level reinforcement learning approach.
\label{appendix:subsec:rl_details}
\paragraph{Exploration.} In practice, we choose large penalty terms $\dist(\cdot,\cdot)$ to minimize greediness. However early in training, randomly initialized penalty terms can harm exploration. We found it was sufficient to down-weight the penalty term early in training. As such, the inner action selection is biased toward short-term rewards, resulting in greedy action selection. However, there are many further possibilities for exploiting random penalty functions to induce exploration, which we discuss in the next section. 

\paragraph{Integer-valued flows.} For several problem settings, it is desirable that the chosen flows be \textit{integer-valued}. For example, in a transportation problem, we may wish to allocate some number of vehicles, which can not be infinitely sub-divided \cite{GammelliYangEtAl2021,GammelliYangEtAl2022}. There are several ways to introduce integer-valued constraints to our framework. First, we note that because the RL agent is trained through policy gradient---and thus we do not require a differentiable inner problem---we can simply introduce integer constraints into the lower-level problem\footnote{Note that several problems exhibit a \textit{total unimodularity} property \cite{Murota2009}, for which the relaxed integer-valued problem is tight. }. However, solving integer-constrained problems is typically expensive in practice. An alternate solution is to simply use a heuristic rounding operation on the output of the inner problem. Again, because of the choice of gradient estimator, this does not need to be differentiable. Moreover, the RL policy learns to adapt to this heuristic clipping. Thus, we in general recommend this strategy as opposed to directly imposing constraints in the inner problem. 

\section{Discussion and Algorithmic Components}
\label{appendix:discussion}

In this section, we discuss various elements of the proposed framework, highlight correspondences and design decisions, and discuss component-level extensions.

\subsection{Distance metric as value function}
\label{appendix:subsec:distance_metric}
The role of the distance metric (and the generated desired next state) is to capture the value of future reward in the greedy one-step inner optimization problem. This is closely related to the value function in dynamic programming and reinforcement learning, which in expectation captures the sum of future rewards for a particular policy. Indeed, under moderate technical assumptions, our linear problem formulation with stochasticity yields convex expected cost-to-go (the negative of the value) \cite{pereira1991multi, dumouchelle2022neur2sp}.

There are several critical differences between our penalty term and a learned value function. First, a value function in a Markovian setting for a given policy is a function solely of state. For example, in the LCP, a value function would depend only on $\st^{t+1}$. In contrast, our value function depends on $\hat{\st}^{t+1}$, which is the output of a policy which takes $\st^t$ as an input. Thus, the penalty term is a function of both the current and desired next state. Given this, the penalty term is better understood as a local approximation of the value function, for which convex optimization is tractable, or as a form of state-action value function with a reduced action space (also referred to as a Q function). 

The second major distinction between the penalty term and a value function is particular to reinforcement learning. Value functions in modern RL are typically learned via minimizing the Bellman residual \cite{SuttonBarto1998}, although there is disagreement on whether this is a desirable objective \cite{fujimoto2022should}. In contrast, our policy is trained directly via gradient descent on the total reward (potentially incorporating value function control variates). Thus, the objective for this penalty method is better aligned with maximizing total reward.

\subsection{Computational efficiency}
\label{appendix:subsec:computational_efficiency}
Consider solving the full nonlinear control problem via direct optimization over a finite horizon ($T$ timesteps), which corresponds to a model predictive control \cite{RawlingsMayne2013} formulation. How many actions must be selected? The number of possible flows for a fully dense graph (worst case) is $N_v (N_v-1)$. In addition to this, there are $\sum_{i\in\mathcal{V}} N_e(i)$ possible exchange actions; if we assume $N_e$ is the same for all nodes, this yields $N_v N_e$ possible actions. Finally, we have $N_c$ commodities. Thus, the worst-case number of actions to select is $T N_c N_v (N_v + N_e - 1)$; it is evident that for even moderate choices of each variable, the complexity of action selection in our problem formulation quickly grows beyond tractability. 

While moderately-sized problems may be tractable within the direct optimization setting, we aim to incorporate the impacts of stochasticity, nonlinearity, and uncertainty, which typically results in non-convexity. The reinforcement learning approach, in addition to being able to improve directly from data, reduces the number of actions required to those for a single step. If we were to directly parameterize the naive policy that outputs flows and exchanges, this would correspond to $N_c N_v (N_v + N_e - 1)$ actions. For even moderate values of $N_c, N_v, N_e$, this can result in millions of actions. It is well-known that reinforcement learning algorithms struggle with high dimensional action spaces \cite{van2020q}, and thus this approach is unlikely to be successful. In contrast, our bi-level formulation requires only $N_c$ actions for the learned policy, while additionally leveraging the beneficial inductive biases over short time horizons.



\section{Additional Experiment Details}
\label{appendix:experiments}
In this section, we provide additional details of the experimental set-up and hyperparameters.
All RL modules were implemented using PyTorch \cite{PaszkeGrossEtAl2019} and the IBM CPLEX solver \cite{ios_ILOG:1987} for the optimization problem.

\subsection{Minimum Cost Flow}
\label{appendix:subsec:minimum_cost_flow}
We start by describing the properties of the environments in Section \ref{appendix:subsubsec:min_cost_flow_environment_details}. 
We further expand the discussion on model implementation (Section \ref{appendix:subsubsec:min_cost_flow_model_implementation}), and additional results (Section \ref{appendix:subsubsec:min_cost_flow_additional_results}).

\subsubsection{Environment details}
\label{appendix:subsubsec:min_cost_flow_environment_details}
We select environment variables in a way to cover a wide enough range of possible scenarios, e.g., different travel times and thus, different optimal actions.
\label{appendix:subsec:environments}

\paragraph{Generalities.} 
As discussed in Section \ref{sec:experiments}, the environments describe a dynamic minimum cost flow problem, whereby the goal is to let commodities flow from source to sink nodes in the minimum time possible (i.e., cost is equal to time). Formally, given a graph $\mathcal{G} = \{\mathcal{V}, \mathcal{E}\}$, the reward function across all environments is defined as:
$$
R(\st^t, \ac^t) = - \sum_{(i,j) \in \mathcal{E}} f_{ij}^t t_{ij} + \lambda f_{\text{sink}}^t,
$$
where $f_{ij}^t$ and $t_{ij}$ represent flow and travel time along edge $(i,j)$ at time $t$, respectively, $f_{\text{sink}}^t$ is the flow arriving at all sink nodes at time $t$, and $\lambda$ is a weighting factor between the two reward terms. In our experiments, the resulting policy proved to be broadly insensitive to values of $\lambda$, with $\lambda \in [15, 30]$ typically being an effective range. 

\paragraph{2-hop, 3-hop, 4-hop.} 
Given a single-source, single-sink network, we assume travel times to be constant over the episode and requirements (i.e., demand) to be sampled at each time step as $\rho = 10 + \psi_i, \psi_i \sim \text{Uniform}[-2, 2]$.
Capacities $c_{ij}$ are fixed to a very high positive number, thus not representing a constraint in practice.
Cost $m_{ij}$ is equal to the travel time $t_{ij}$.
An episode is assumed to have a duration of 30 time steps and terminates when there is no more flow traversing the network. 
To present a variety of scenarios to the agent at training time, we sample random travel times for each new episode as $t_{ij} \sim \text{Uniform}[0, 10]$ and use the topologies shown in Fig. \ref{fig:topologies}.
In our experiments, we apply as many layers of message passing as hops from source to sink node in the graph, e.g., $K=2$ and $K=3$ in the 2-hops and 3-hops environment, respectively.

\paragraph{Dynamic travel times}
To train our MPNN-RL, we select the 3-hops environment and generate travel times as follows for every episode: (i) sample random travel times as $t_{ij} \sim \text{Uniform}[0, 10]$, (ii) for every time step, gradually change the travel time as $t_{ij} = t_{ij} + \psi, \psi \sim \text{Uniform}[-1,1]$.


\paragraph{Capacity constraints.} 
In this experiment, we focus on the 3-hops environment and assume a constant value $c_{ij} = 20, \forall (i,j) \in \edges: j \neq 7$ while we keep a high value for all the edges going into node $7$ (i.e., the sink node) which would more easily generate infeasible scenarios.
From an RL perspective, we add the following edge-level features:
\begin{itemize}
    \item Edge-capacity $\{c_{ij}^t\}_{i,j)\in\edges}$ at the current time step $t$.
    \item Accumulated flow $\{f_{ij}^t\}_{(i,j)\in\edges}$ on edge $(i,j)$
\end{itemize}

\paragraph{Multi-commodity.} 
Let $N_c$ define the number of commodities to consider, indexed by $k$.
From an RL perspective, we extend the proposed policy to represent a $N_c$-dimensionsional Dirichlet distribution.
Concretely, we define the output of the policy network to represent the $N_c \times N_v$ concentration parameters $\mathbf{\alpha} \in \mathbb{R}_+^{N_c \times N_v}$ of a Dirichlet distribution over nodes for each commodity, such that $\mathbf{a}_t \sim \text{Dir}\{\mathbf{a}_t | \mathbf{\alpha}\}$.
In other words, to extend our approach to the multi-commodity setting, we define a multi-head policy network characterized by one head per commodity.
In our experiments, we train our multi-head agent on the topology shown in Fig. \ref{fig:multi_commodity} whereby we assume two parallel commodities: commodity A going from node 0 to node 10, and commodity B going from node 0 to node 11.
We choose this topology so that the only way to solve the scenario is to discover distinct behaviours between the two network heads (i.e., the policy head controlling flow for commodity A needs to go up or it won't get any reward, and vice-versa for commodity B).

\paragraph{Computational analysis.}
In this experiment, we generate different versions of the 3-hops environment, whereby different environments are characterized by intermediate layers with increasing number of nodes and edges.
The results are computed by applying the pre-trained MPNN-RL agent on the original 3-hops environment (i.e., characterized by 8 nodes in the graph). 
In light of this, Figure \ref{fig:computational_analysis} showcases a promising degree of transfer and generalization among graphs of different dimensions.

\subsubsection{Model implementation}
\label{appendix:subsubsec:min_cost_flow_model_implementation}
In our experiments, we implement the following methods:

\paragraph{Randomized heuristics.} 
In this class of methods, we focus on measuring performance of simple heuristics.
\begin{enumerate}
    \item \emph{Random policy}: at each timestep, we sample the desired next state from a Dirichlet prior with concentration parameter $\alpha = [1, 1, \ldots, 1]$. 
    This benchmark provides a lower bound of performance by choosing desired next states randomly.
\end{enumerate}

\paragraph{Learning-based.} Within this class of methods, we focus on measuring how different architectures affect the quality of the solutions for the dynamic network control problem.
For all methods, the A2C algorithm is kept fixed, thus the difference solely lies in the neural network architecture.
    \begin{enumerate}
        \setcounter{enumi}{1}
        \item \emph{MLP-RL}: both policy and value function estimator are parametrized by feed-forward neural networks. In all our experiments, we use two layers of 32 hidden unites and an output layer mapping to the output's support (e.g., a scalar value for the critic network). Through this comparison, we highlight the performance and flexibility of graph representations for network-structured data.
        \item \emph{GCN-RL}: In all our experiments, we use $K$ layers of graph convolution with 32 hidden units, with $K$ equal to the number of sink-to-source hops in the graph, and a linear output layer mapping to the output's support. See below for a broader discussion of graph convolution operators.
        \item \emph{GAT-RL}: In all our experiments, we use $K$ layers of graph attention \cite{Velickovic2018} with 32 hidden units, with $K$ equal to the number of sink-to-source hops in the graph, and single attention head. The output is further computed by a linear output layer mapping to the output's support. Together with GCN-RL, this model represents an approach based on graph convolutions rather than explicit message passing along the edges (as in MPNNs). Through this comparison, we argue in favor of explicit, pair-wise messages along the edges, opposed to sole aggregation of node features among a neighborhood. Specifically, we argue in favor of the alignment between MPNN and the kind of computations required to solve flow optimization tasks, e.g., propagation of travel times and selection of best path among a set of candidates (max aggregation). 
        \item \emph{MPNN-RL}: ours. We use $K$ layers of message passing neural network \cite{GilmerEtAl2017} of 32 hidden units as defined in Section \ref{appendix:subsec:network_architeture}, with $K$ equal to the number of sink-to-source hops in the graph, and a linear output layer mapping to the output's support.
    \end{enumerate}

\noindent \textbf{MPC-based.} Within this class of methods, we focus on measuring performance of MPC approaches that serve as state-of-art benchmarks for the dynamic network flow problem. 

\begin{enumerate}
    \setcounter{enumi}{5}
    \item \emph{Oracle}: we directly optimize the flow using a standard formulation of MPC \cite{ZhangRossiEtAl2016b}. Notice that although the embedded optimization is a linear programming model, it may not meet the computation requirement of real-time applications (e.g., obtaining a solution within several seconds) for large scale networks. 
    In this work, MPC is assumed to have access to future state elements (e.g., future travel times, connectivity, etc.). Crucially, assuming knowledge of future state elements is equivalent to assuming oracle knowledge of the realization of all stochastic elements in the system. In other words, there is no uncertainty for the MPC (this is in contrast with RL-based benchmarks, that assume access only to \textit{current} state elements).
    In our experiments, the benchmark with the ``Oracle" MPC enables us to quantify the optimal solution for all environments, thus giving a sense of the optimality gap between the ground truth optimum and the solution achieved via RL.
\end{enumerate}

\subsubsection{Additional results}
\label{appendix:subsubsec:min_cost_flow_additional_results}
\begin{figure}[th]
      \centering
     \includegraphics[width=\textwidth]{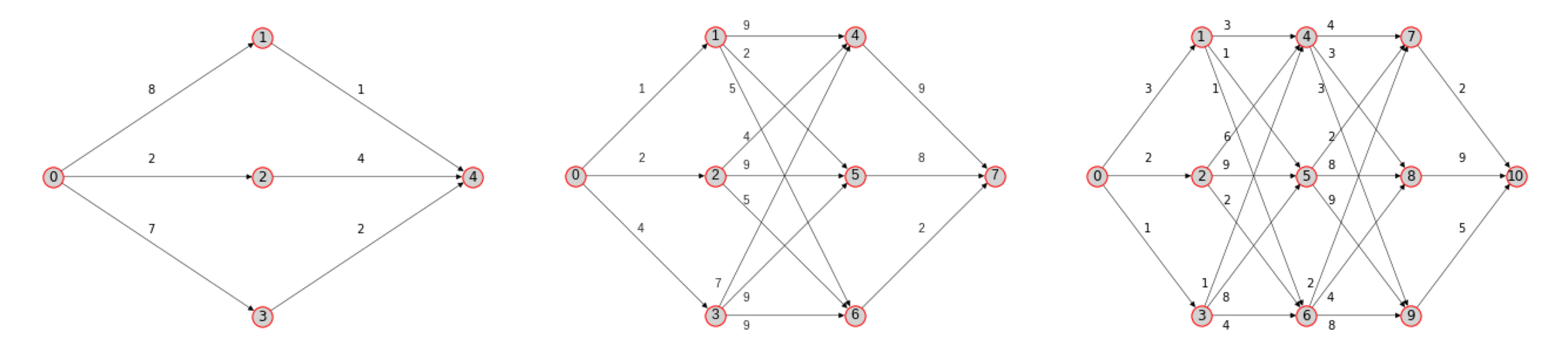}
      \caption{Graph topologies used for the message passing experiments: 2-hops (left), 3-hops (center), 4-hops (right). The source and sink nodes are represented by the left-most and right-most nodes, respectively. Values in the proximity of the edges represent travel times.}
      \label{fig:topologies}
\end{figure}

\paragraph{Minimum cost flow through message passing.}
In this first experiment, we consider 3 different environments (Fig. \ref{fig:topologies}), such that different topologies enforce a different number of required hops of message passing between source and sink nodes to select the best path.
Results in Table \ref{tab:experiments} (\textit{2-hop}, \textit{3-hop}, \textit{4-hop}) show how MPNN-RL is able to achieve at least $87\%$ of oracle performance.
Table \ref{tab:experiments} further shows how agents based on graph convolutions (i.e., GCN, GAT) fail to learn an effective flow optimization strategy. 
   
\begin{figure}[th]
      \centering
     \includegraphics[width=\textwidth]{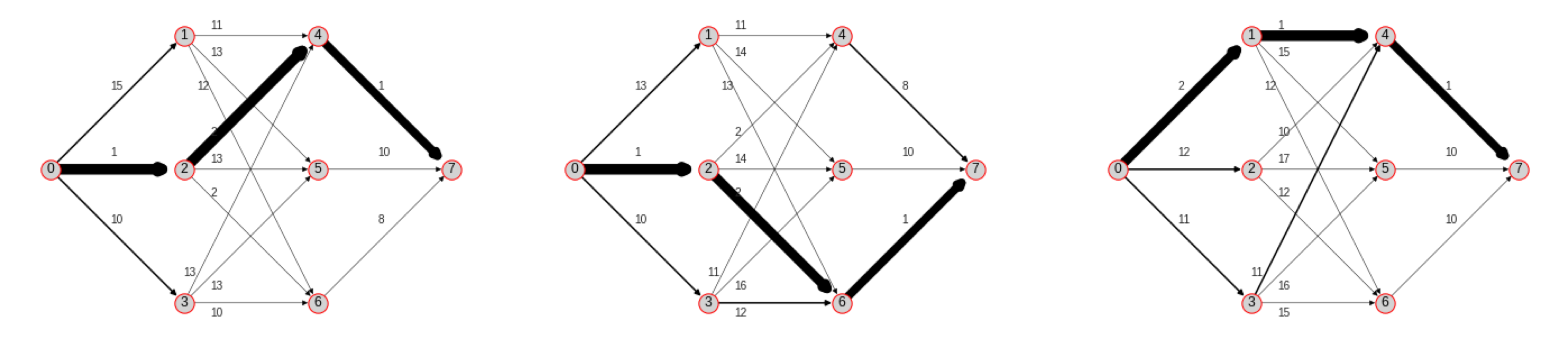}
      \caption{Visualization of a trained instance of MPNN-RL on an environment with dynamic travel times. We simulate a scenario where the optimal path changes three times (left, middle, and right) over the course of an episode. Shaded edges represent actions induced by the MPNN-RL.}
      \label{fig:dynamic_tt}
\end{figure}

\paragraph{Dynamic travel times.}
In many real-world systems, travel times evolve over time.
To approach this, in Fig. \ref{fig:dynamic_tt} and Table \ref{tab:experiments} (\textit{Dyn travel time}) we measure results on a dynamic network characterized by two \textit{change-points}, i.e., time steps where the optimal path changes because of a change in travel times.
Results show how the proposed MPNN-RL is able to achieve above $99\%$ of oracle performance.

\begin{figure}[th]
      \centering
     \includegraphics[width=\textwidth]{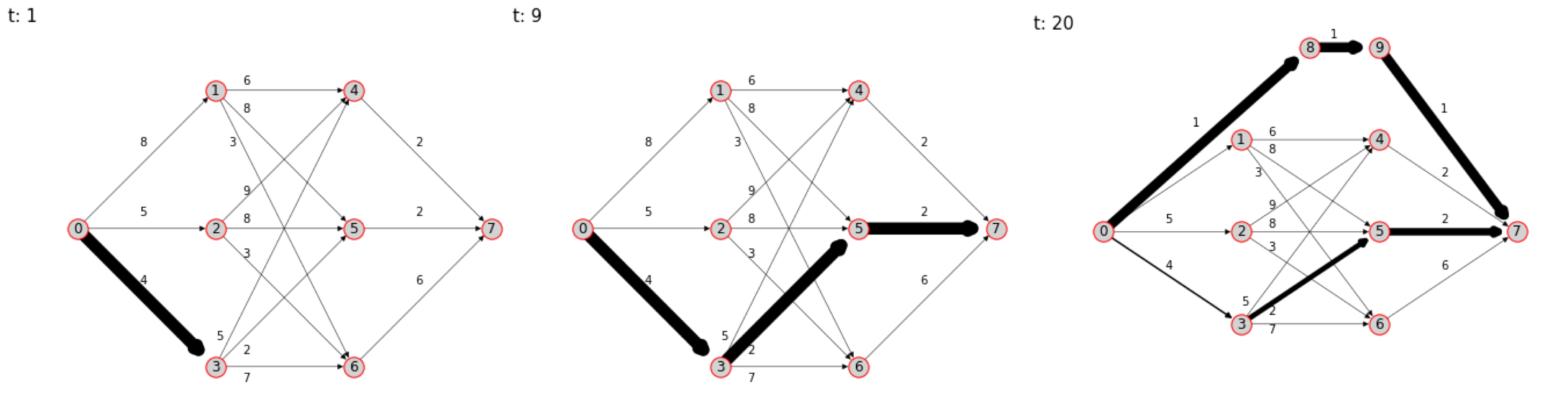}
      \caption{Visualization of a trained instance of MPNN-RL on an environment with dynamic topology. We simulate a scenario where the optimal path changes over the course of an episode because of the addition of a new path. Shaded edges represent actions induced by the MPNN-RL.}
      \label{fig:dynamic_topology}
   \end{figure}

\paragraph{Dynamic topology.}
In real-world systems, operations are often characterized by time-dependent topologies, i.e., nodes and edges can be dropped or added during an episode, such as in roadblocks within transportation systems or the opening of a new shipping center in supply chain networks.
However, most traditional approaches cannot deal with these conditions easily.
On the other hand, the locality of graph network-based agents, together with the one-step implicit planning of RL, enable our framework to deal with multiple time-varying graph configurations during the same episode.
Fig. \ref{fig:dynamic_topology} and Table \ref{tab:experiments} (\textit{Dyn topology}) show how MPNN-RL achieves 83.9\% of oracle performance clearly outperforming the other benchmarks.
Crucially, these results highlight how agents based on MLPs result in highly inflexible network controllers that are limited to the same topology they were exposed to during training.

\begin{figure}[th]
      \centering
     \includegraphics[width=1\textwidth]{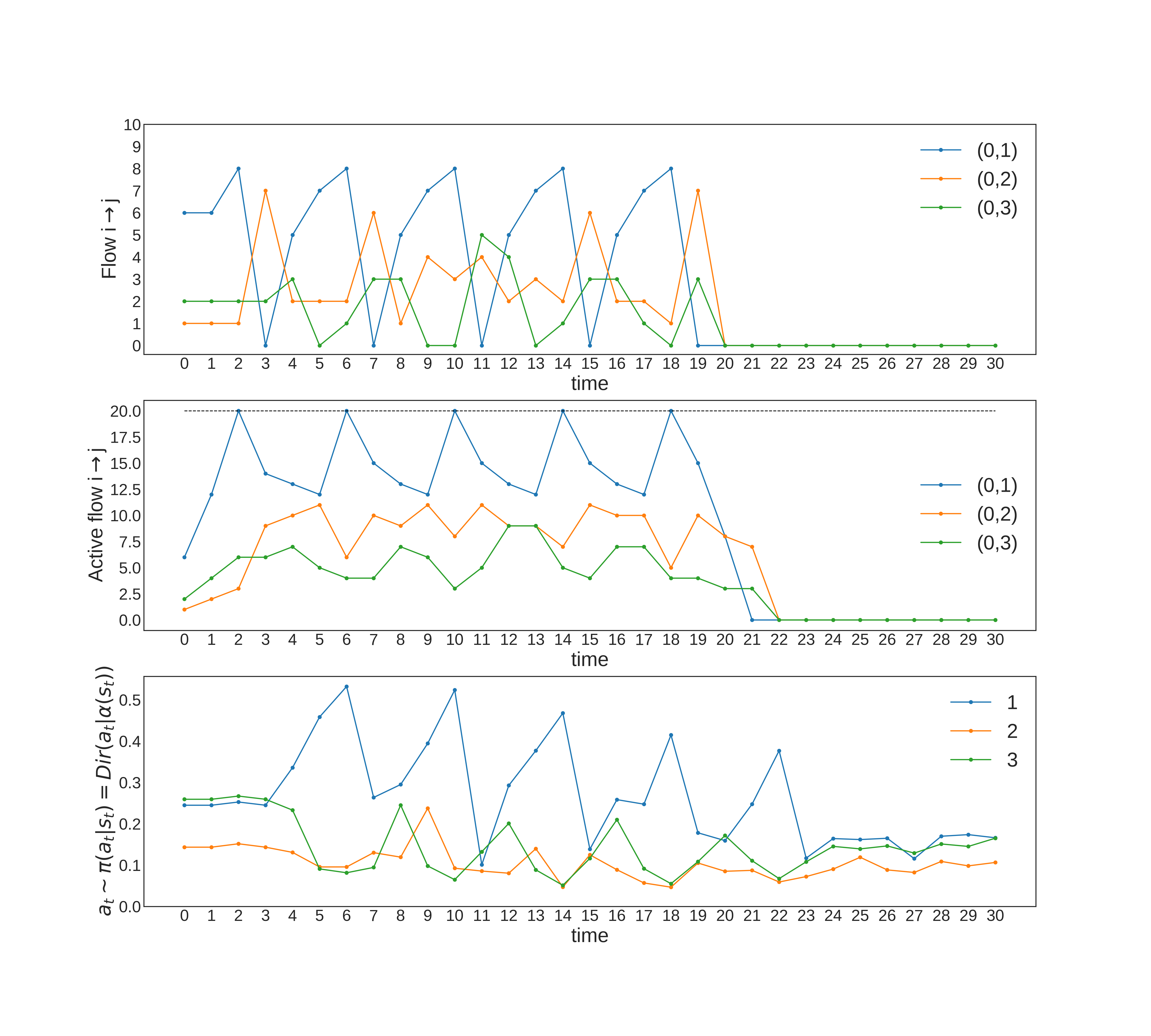}
      \caption{Visualization of the MPNN-RL policy on the capacity-constrained environment. (Top) The resulting flow $f_{ij}$ on the edges $0\rightarrow1, 0\rightarrow2, 0\rightarrow3$. (Center) The accumulated flow on the same edges compared to the fixed capacity $c_{ij} = 20$, represented as a dashed horizontal line. (Bottom) The desired distribution described by the MPNN-RL policy.}
      \label{fig:capacity_constrained}
\end{figure}

\paragraph{Capacity constraints.}
Real-world systems are often represented as capacity-constrained networks.
In this experiment, we relax the assumption that capacities $c_{ij}$ are always able to accommodate any flow on the graph. 
Compared to previous sections, the lower capacities introduce the possibility of infeasible states.
To measure this, the \textit{Success Rate} computes the percentage of episodes which have been terminated successfully.
Results in Table \ref{tab:experiments} (\textit{Capacity}) highlight how MPNN-RL is able to achieve 89.8\% of oracle performance while being able to successfully terminate 87\% of episodes.
Qualitatively, Fig. \ref{fig:capacity_constrained} shows a visualization of the policy for a specific test episode. 
The plots show how MPNN-RL is able to learn the effects of capacity on the optimal strategy by allocating flow to a different node when the corresponding edge is approaching its capacity limit.
   
\begin{figure}[th]
      \centering
     \includegraphics[width=\textwidth]{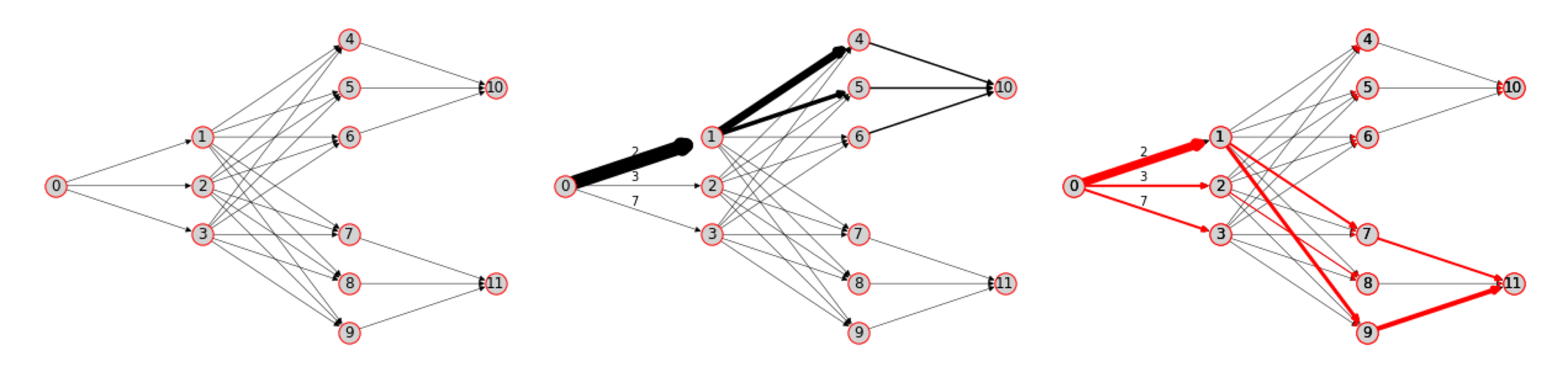}
      \caption{Visualization of the multi-commodity environment. (Left) The topology considered during our experiments. (Center) A visualization of the policy for the first commodity A. (Right) A visualization of the policy for the second commodity B.}
      \label{fig:multi_commodity}
\end{figure}

\paragraph{Multi-commodity.}
Often, system operators might be interested in controlling multiple commodities over the same network.
In this scenario, we extend the current architecture to deal with multiple commodities and source-sink combinations.
Results in Table \ref{tab:experiments} (\textit{Multi-commodity}) and Fig. \ref{fig:multi_commodity} show how MPNN-RL is able to effectively recover distinct policies for each commodity, thus being able to operate successfully multi-commodity flows within the same network.

\subsection{Supply Chain Inventory Management}
\label{appendix:subsec:scim}
We start by describing the properties of the environments in Section \ref{appendix:subsubsec:scim_environment_details}. 
We further expand the discussion on MDP definitions (Section \ref{appendix:subsubsec:scim_mdp_details}), model implementation (Section \ref{appendix:subsubsec:scim_model_implementation}), and specifics on the linear control problem (Section \ref{appendix:subsubsec:scim_lcp}).

\subsubsection{Environment details}
\label{appendix:subsubsec:scim_environment_details}
In our experiments, all stores are assumed to have an independent demand-generating process. 
We simulate a seasonal demand behavior by representing the demand as a co-sinusoidal function with a stochastic component, defined as follows:
\begin{equation}
    \demand_i^t = \left\lfloor \frac{d_i^{\max}}{2} \left(1 + \cos \left(\frac{4\pi(2i + t)}{T}\right)\right) + \mathcal{U}(0, d_i^{var}) \right\rfloor,
\end{equation}
where $\lfloor \cdot \rfloor$ is the floor function, $d_i^{\max}$ is the maximum demand value, $\mathcal{U}(0, d_i^{var})$ is a uniformly distributed random variable, and $T$ is the episode length.

Environment parameters are defined as follows:
\begin{table}[h]
\centering
\small
\caption{Parameters for the 1F2S environment}
\begin{tabular}{l l p{4.5cm} l l p{3cm}}
    Parameter & Explanation & Value & Parameter & Explanation & Value \\
    \midrule 
    $d^{\max}$ & Maximum demand & [2, 16] & $m^S$ & Storage cost & [3, 2, 1] \\ [0.1ex]
    $d^{var}$ & Demand variance & [2, 2] & $m^O$ & Production cost & 5 \\ [0.1ex]
    $T$ & Episode length & 30 & $m^B$ & Backorder cost & 21 \\ [0.1ex]
    $t^P$ & Production time & 1 & $m^T$ & Transportation cost & [0.3, 0.6] \\ [0.1ex]
    $t_{ij}$ & Travel time & [1, 1] & $p$ & Price & 15 \\ [0.1ex]
    $c$ & Storage capacity & [20, 9, 12] & & &\\ [0.1ex]
    \bottomrule
    \end{tabular}%
\label{tab:scim_parameters_1f2s}%
\end{table}

\begin{table}[h]
\centering
\small
\caption{Parameters for the 1F3S environment}
\begin{tabular}{l l p{4.5cm} l l p{3cm}}
    Parameter & Explanation & Value & Parameter & Explanation & Value \\
    \midrule 
    $d^{\max}$ & Maximum demand & [1, 5, 24] & $m^S$ & Storage cost & [2, 1, 1] \\ [0.1ex]
    $d^{var}$ & Demand variance & [2, 2, 2] & $m^O$ & Production cost & 5 \\ [0.1ex]
    $T$ & Episode length & 30 & $m^B$ & Backorder cost & 21 \\ [0.1ex]
    $t^P$ & Production time & 1 & $m^T$ & Transportation cost & [0.3, 0.3, 0.3] \\ [0.1ex]
    $t_{ij}$ & Travel time & [1, 1, 1] & $p$ & Price & 15 \\ [0.1ex]
    $c$ & Storage capacity & [30, 15, 15, 15] & & &\\ [0.1ex]
    \bottomrule
    \end{tabular}%
\label{tab:scim_parameters_1f3s}%
\end{table}

\begin{table}[h]
\centering
\small
\caption{Parameters for the 1F10S environment}
\begin{tabular}{l l p{4.5cm} l l p{3cm}}
    Parameter & Explanation & Value & Parameter & Explanation & Value \\
    \midrule 
    $d^{\max}$ & Maximum demand & [2, 2, 2, 2, 10, 10, 10, 18, 18, 18] & $m^S$ & Storage cost & $[1, 2 \quad \forall i \in \nodes / 0]$ \\ [0.1ex]
    $d^{var}$ & Demand variance & $[2]_{i \in \nodes}$ & $m^O$ & Production cost & 5 \\ [0.1ex]
    $T$ & Episode length & 30 & $m^B$ & Backorder cost & 21 \\ [0.1ex]
    $t^P$ & Production time & 1 & $m^T$ & Transportation cost & $[0.3]_{i \in \nodes}$ \\ [0.1ex]
    $t_{ij}$ & Travel time & $[1]_{i \in \nodes}$ & $p$ & Price & 15 \\ [0.1ex]
    $c$ & Storage capacity & $[100, 15 \quad \forall i \in \nodes / 0]$ & & &\\ [0.1ex]
    \bottomrule
    \end{tabular}%
\label{tab:scim_parameters_1f10s}%
\end{table}

\subsubsection{MDP details}
\label{appendix:subsubsec:scim_mdp_details}
In what follows, we complement Section \ref{subsec:supply_chain_inventory_management} with a formal definition of the SCIM MDP.

\textit{Reward $\left(R(\st^t, \ac^t)\right)$:} we select the reward function in the MDP as the profit of the inventory manager, computed as the difference between revenues and the sum of storage, production, transportation, and backorder costs: 
\begin{align}
    R(\st^t, \ac^t) & = \sum_{i \in \warehousenodes} p \cdot \min(\demand_i^t, \inventory_i^t) - \left(\sum_{i \in \nodes} \storagecost_i \cdot \inventory_i^t + \sum_{i \in \warehousenodes} \ordercost_i \cdot \order_i^t + \sum_{(i,j) \in \edges} \transportcost_{ij} \cdot \flow_{ij}^t - \sum_{i \in \storenodes} \backordercost_i \cdot \min(0, \inventory_i^t) \right).
\end{align}

\textit{State space ($\stspace$):} the state space contains information to describe the current status of the supply network, via the definition of node and edge features.
Node features contain information on (i) current inventory $\inventory_i$, (ii) current and estimated demand for the next $T$ timesteps $\hat \demand_i^{t:t+T}$, (iii) incoming flow for the next $T$ timesteps $\sum_{j \in \nodes} \flow_{ji}^{t:t+T}$, and (iv) incoming orders for the next $T$ timesteps $\order_i^{t:t+T}$, such that $\bx_i = [\inventory_i, \hat \demand_i^{t:t+T}, \sum_{j \in \nodes} \flow_{ji}^{t:t+T}, \order_i^{t:t+T}]$.
Edge features are represented by the concatenation of (i) travel time $t_{ij}$, and (ii) transportation cost $\transportcost_{ij}$, such that $\be_{ij} = [t_{ij}, \transportcost_{ij}]$.

\subsubsection{Model implementation}
\label{appendix:subsubsec:scim_model_implementation}
\begin{figure}[th]
    \centering
    \subfigure[]{\includegraphics[width=0.33\textwidth]{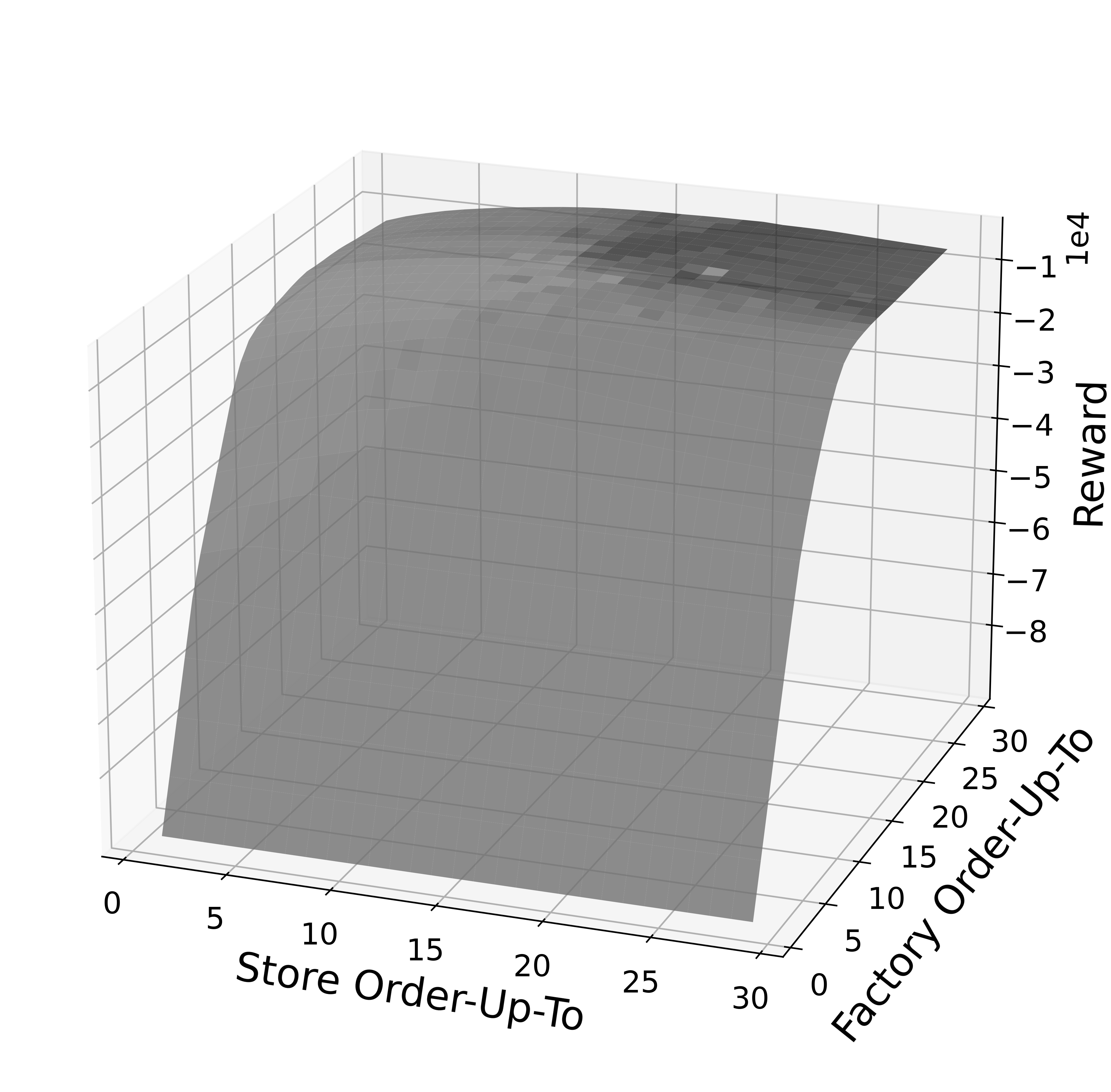}} 
    \subfigure[]{\includegraphics[width=0.33\textwidth]{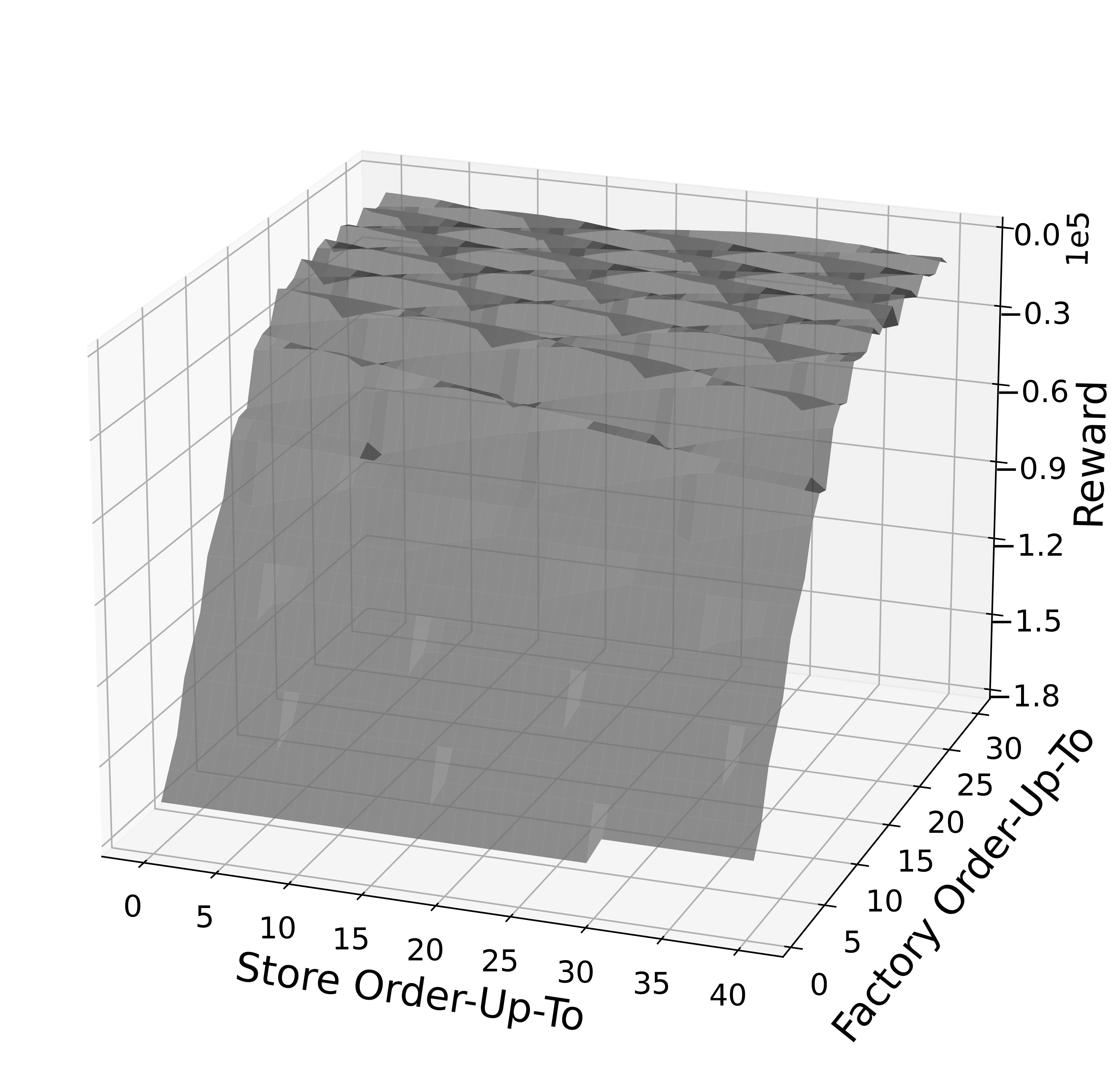}} 
    \subfigure[]{\includegraphics[width=0.33\textwidth]{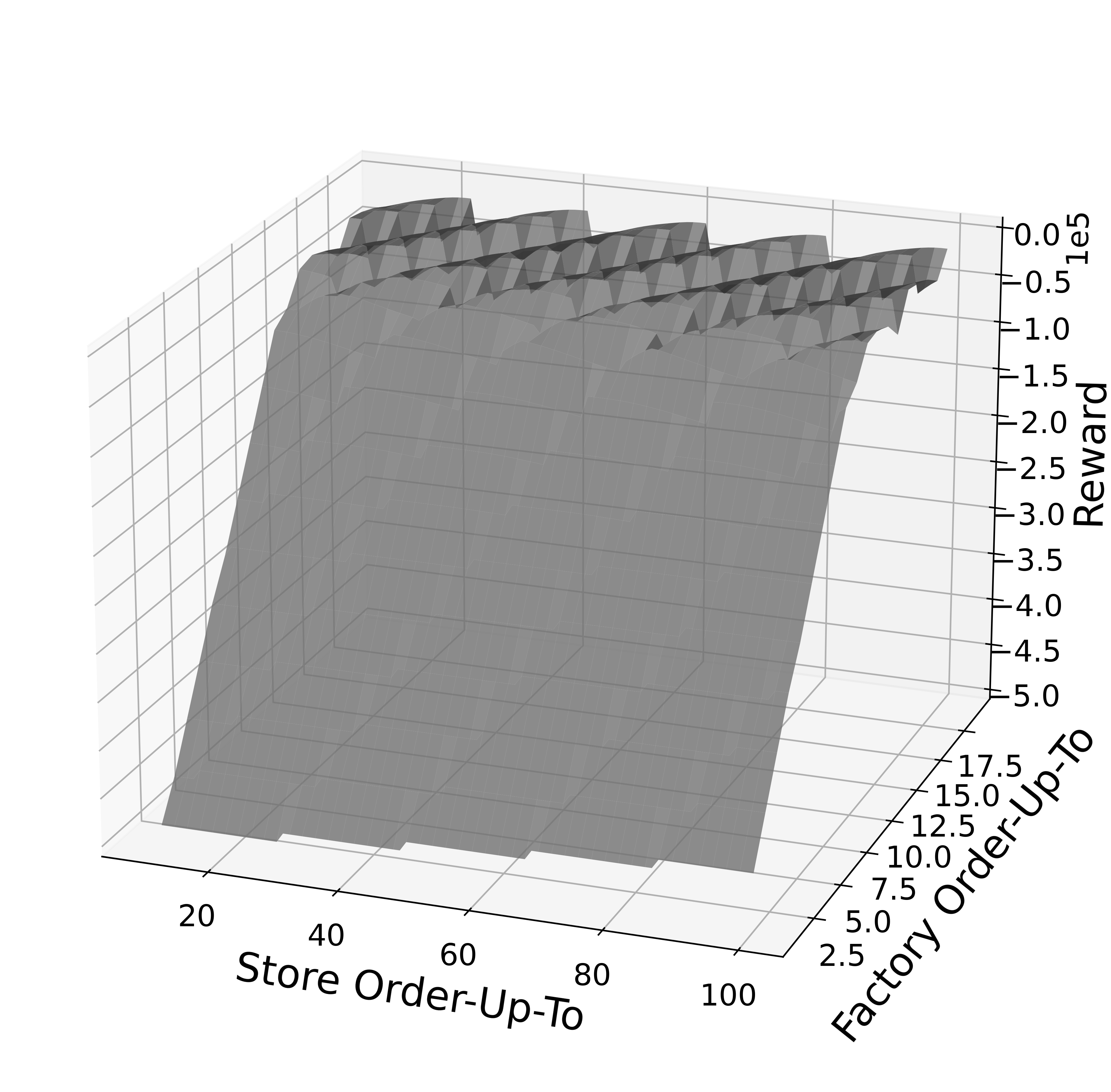}}
    \caption{Parameter tuning for the S-type policy on (a) 1F2S, (b) 1F3S, and (c) 1F10S environments.}
    \label{fig:s_type}
\end{figure}

In what follows, we provide additional details for the implemented baselines and models:
\paragraph{Randomized heuristics.} 
In this class of methods, we focus on measuring the performance of simple heuristics.
\begin{enumerate}
    \item \emph{Avg. Prod}: at each timestep, we (1) select production $w_i^t$ to be the average episode demand across all stores, and (2) sample the desired distribution from a Dirichlet prior with concentration parameter $\alpha = [1, 1, \ldots, 1]$ to simulate a random shipping behavior. 
\end{enumerate}
\paragraph{Domain-driven heuristics.}
Within this class of methods we measure the performance of heuristics generally accepted as effective baselines.
\begin{enumerate}
    \setcounter{enumi}{1}
    \item \emph{S-type Policy}: also referred to as ``order-up-to'' policy, this heuristic is parametrized by two values: a warehouse order-up-to level and a store order-up-to level.
    At a high level, at each time step the inventory manager aims to order inventory such that all inventory at and expected to arrive at the warehouse and at the stores is equal to the warehouse order-up-to level and the store order-up-to level, respectively. 
    Concretely, we fine-tune the S-type policy on each environment individually by running an exhaustive search for the best order-up-to levels, as shown in Figure \ref{fig:s_type}.
\end{enumerate}
\paragraph{Learning-based approaches.} 
    \begin{enumerate}
        \setcounter{enumi}{2}
        \item \emph{End-to-end RL}: with this benchmark, we evaluate the performance of RL architectures that do not approach the problem via the proposed bi-level formulation. Specifically, as traditionally done in RL, we define the policy network to represent a direct mapping from states to environment actions. In our experiments, both policy and value function estimator are parametrized by feed-forward neural networks with two layers of 64 hidden units followed by linear layers mapping to either (i) mean and standard deviation parameters for the policy network, or (ii) a scalar value function estimate for the critic. Among the three scenarios, we adjust the input layer based on the input dimensionality (which is topology-dependent since we unroll all node and edge features into a vector representation of the graph). Through this comparison, we highlight the benefits of the bi-level formulation for graph control problems.
        \item \emph{Graph-RL}: ours. We use $K=2$ layers of message passing neural network \cite{GilmerEtAl2017} of 32 hidden units with sum aggregation function as defined in Section \ref{appendix:subsec:network_architeture} followed by a linear layer mapping to the output's support.
    \end{enumerate}

\subsubsection{LCP formulation}
\label{appendix:subsubsec:scim_lcp}
Given a desired next state described by (i) the desired production in warehouse nodes $\hat \order_i^{t+1}, \forall i \in \warehousenodes$, and (ii) a desired inventory in store nodes $\hat \inventory_i^{t+1}, \forall i \in \storenodes$, we define the following linear control problem as follows:
\begin{subequations}
\begin{align}
    \min_{\flow_{ij}^t, \order^t, \epsilon_{i}^{\flow}, \epsilon_{i}^{\order}} & \sum_{i \in \storenodes} |\epsilon_{i}^\flow| + \sum_{i \in \warehousenodes} |\epsilon_i^\order| & \label{eq:scim_lcp_obj}\\
    \textrm{s.t.} & \sum_{j\in \mathcal{N}^-(i)} \flow_{ji}^t = \hat \inventory_i^{t+1} + \epsilon_i^\flow, & i\in\storenodes \label{eq:scim_lcp_con1}\\
    & \inventory_i^t + \sum_{j\in \mathcal{N}^-(i)} \flow_{ji}^t - \demand_i^t \leq \capacity_i^t, & i\in\storenodes \label{eq:scim_lcp_con2}\\
    & \sum_{j \in \mathcal{N}^+(i)} \flow_{ij}^t \leq \inventory_i^t, & i \in \warehousenodes \label{eq:scim_lcp_con3}\\
    & \inventory_i^t + \order_i^t - \sum_{j \in \mathcal{N}^+(i)} \flow_{ij}^t \leq \capacity_i^t, & i \in \warehousenodes \label{eq:scim_lcp_con4}\\
    & \order_i^t = \hat \order_i^t + \epsilon_i^\order, & i \in \warehousenodes \label{eq:scim_lcp_con5}\\
    & \flow_{ij}^t \geq 0, & (i,j) \in \edges \label{eq:scim_lcp_con6}
\end{align}\label{eq:scim_lcp}
\end{subequations}
where, as introduced in Section \ref{sec:methodology}, the objective function \eqref{eq:scim_lcp_obj} represents the distance metric $d(\cdot, \cdot)$ that penalizes the deviation from the desired next states, the constraint \eqref{eq:scim_lcp_con1} ensures that the total incoming flow in store nodes is as close as possible to the desired inventory, the constraint \eqref{eq:scim_lcp_con2} represents that the inventory in store nodes after shipping and demand satisfaction does not exceed storage capacity, the constraint \eqref{eq:scim_lcp_con3} ensures that the shipped products are upper bounded by inventory, constraint \eqref{eq:scim_lcp_con4} represents that the inventory in warehouse nodes after shipping and re-ordering does not exceed storage capacity, constraint \eqref{eq:scim_lcp_con5} ensures that orders from manufacturers are close to the desired orders specified through RL, and lastly that commodity flows are non-negative via \eqref{eq:scim_lcp_con6}.

\subsection{Dynamic Vehicle Routing}
\label{appendix:subsec:dvr}
We start by describing the properties of the environments in Section \ref{appendix:subsubsec:dvr_environment_details}. 
We further expand the discussion on MDP definition (Section \ref{appendix:subsubsec:dvr_mdp_details}), model implementation (Section \ref{appendix:subsubsec:dvr_model_implementation}), specifics on the linear control problem (Section \ref{appendix:subsubsec:dvr_lcp}), and additional results (Section \ref{appendix:subsubsec:dvr_additional_results}).

\subsubsection{Environment details}
\label{appendix:subsubsec:dvr_environment_details}
We use two case studies from the cities of New York, USA, and Shenzhen, China, whereby we study a hypothetical deployment of taxi-like systems to serve the peak-time commute demand in popular areas of Brooklyn and Shenzhen, respectively. 
The cities are divided into geographical areas, each of which represents a station. 
The case studies in our experiments are generated using trip record datasets, which we provide together with our codebase.
The trip records are converted to demand, travel times, and trip prices between stations. 
Here, we consider stochastic time-varying demand patterns, whereby customer arrival is assumed to be a time-dependent Poisson process, and the Poisson rates are aggregated from the trip record data every 3 minutes. 
We assume the stations to be spatially connected, whereby moving vehicles from one station to the other requires non-trivial sequential actions (i.e., vehicles cannot directly be repositioned from one station to any other station, rather they have to adhere to the available paths given by the city's topology).

The following remarks are made in order. First, we assume travel times are given and independent of operator actions. This assumption applies to cities where the number of vehicles in the fleet constitutes a relatively small proportion of the entire vehicle population on the transportation network, and thus the impact on traffic congestion is marginal. This assumption can be relaxed by training the proposed RL model in an environment considering the endogenous congestion caused by controlled vehicles fleet. Second, without loss of generality, we assume that the arrival process of passengers for each origin-destination pair is a time-dependent Poisson process. We further assume that such process is independent of the arrival processes of other origin-destination pairs and the coordination of vehicles. These assumptions are commonly used to model transportation requests \cite{Daganzo1978}.

\subsubsection{MDP details}
\label{appendix:subsubsec:dvr_mdp_details}
In what follows, we complement Section \ref{subsec:dynamic_vehicle_routing} with a formal definition of the SCIM MDP.

\textit{Reward $\left(R(\st^t, \ac^t)\right)$:} we select the reward function in the MDP as the operator profit, computed as the difference between revenues and operation-related costs: 
\begin{equation}
    R(\st^t, \ac^t) = \sum_{(i,j) \in \edges} \flow_{ij,P}^t \cdot (\price_{ij} - m_{ij}) - \sum_{(i,j) \in \edges} \flow_{ij,R}^t  \cdot m_{ij}
    \label{eq:dvr_reward}
\end{equation}

\textit{State space ($\stspace$):} the state space contains information to describe the current status transportation network via the definition of node features.
Node features contain information on (i) the current availability of idle vehicles in each station $\inventory_i$, (ii) current and estimated demand for the next $T$ timesteps $\hat \demand_i^{t:t+T}$, (iii) projected availability for the next $T$ timesteps $\hat \inventory_i^{t:t+T}$, and (iv) provider-level information such as trip price $p_{ij}$ and cost $z_{ij}$.

\subsubsection{Model implementation}
\label{appendix:subsubsec:dvr_model_implementation}
In what follows, we provide additional details for the implemented baselines and models:

\paragraph{Randomized heuristics.} 
In this class of methods, we focus on measuring the performance of simple heuristics.
\begin{enumerate}
    \item \emph{Random}: at each timestep, we sample the desired distribution from a Dirichlet prior with concentration parameter $\alpha = [1, 1, \ldots, 1]$. 
This benchmark provides a lower bound of performance by choosing desired next states randomly.
\end{enumerate}

\paragraph{Domain-driven heuristics.}
Within this class of methods, we measure the performance of heuristics generally accepted as reasonable baselines.
\begin{enumerate}
    \setcounter{enumi}{1}
    \item \emph{Equally-balanced System}: at each decision, we take rebalancing actions so to recover an equal distribution of idle vehicles across all areas in the transportation network. Concretely, the heuristic achieves this by solving the DVR LCP with a fixed desired number of idle vehicles among all stations, i.e., given $M$ available vehicles at time $t$, $\hat \bq^{t+1} = \{\hat \inventory_i^{t+1}\}_{i \in \nodes} = \{\frac{M}{|\nodes|}\}_{i \in \nodes}$.
\end{enumerate}

\paragraph{Learning-based approaches.}
\begin{enumerate}
        \setcounter{enumi}{2}
        \item \emph{End-to-end RL}: both policy and value function estimator are parametrized by neural networks that mirror the architecture of Graph-RL. While the critic has the exact same architecture, the actor differs in the last layer, which is characterized by an edge convolution (consisting of 2 linear layers of 32 hidden units) that outputs mean and standard deviation parameters of a Gaussian policy for each edge in the graph.
        \item \emph{Graph-RL}: ours. For both actor and critic networks, we use one layer of graph convolution \cite{KipfWelling2017} with 32 hidden units with sum aggregation function as defined in Section \ref{appendix:subsec:network_architeture} followed by 2 linear layers of 32 hidden units and a final linear layer mapping to the respective output's support.
    \end{enumerate}

\subsubsection{LCP formulation}
\label{appendix:subsubsec:dvr_lcp}
Given a desired next state described by the desired number of idle vehicles across stations $\hat \inventory_i^{t+1}, \forall i \in \nodes$, we define the following linear control problem as follows:
\begin{subequations}
\begin{align}
    \min_{\flow_{ij,R}^t} & \sum_{(i,j)\in \edges}m_{ij}^t \flow_{ij,R}^t & \label{eq:dvr_obj}\\
    \textrm{s.t.} &
    \sum_{j\neq i}(\flow_{ji,R}^t - \flow_{ij,R}^t) + \inventory_i^t \geq \hat \inventory_i^t , & i\in\nodes\label{eq:dvr_con1} \\
    & \sum_{j\neq i} \flow_{ij,R}^t \leq \inventory_i^t, & i\in\nodes\label{eq:dvr_con2} \\
    & \flow_{ij,R}^t \geq 0, & (i,j) \in \edges \label{eq:dvr_con3} 
\end{align}
\label{eq:dvr}
\end{subequations}
where the objective function \eqref{eq:dvr_obj} represents the rebalancing cost, constraint \eqref{eq:dvr_con1}  ensures that the resulting number of vehicles is close to the desired number of vehicles, and with constraints \eqref{eq:dvr_con2}, \eqref{eq:dvr_con3} ensuring that the total rebalancing flow from a region is upper-bounded by the number of idle vehicles in that region and non-negative.

\subsubsection{Additional results}
\label{appendix:subsubsec:dvr_additional_results}
\begin{figure}[th]
      \centering
     \includegraphics[width=0.6\textwidth]{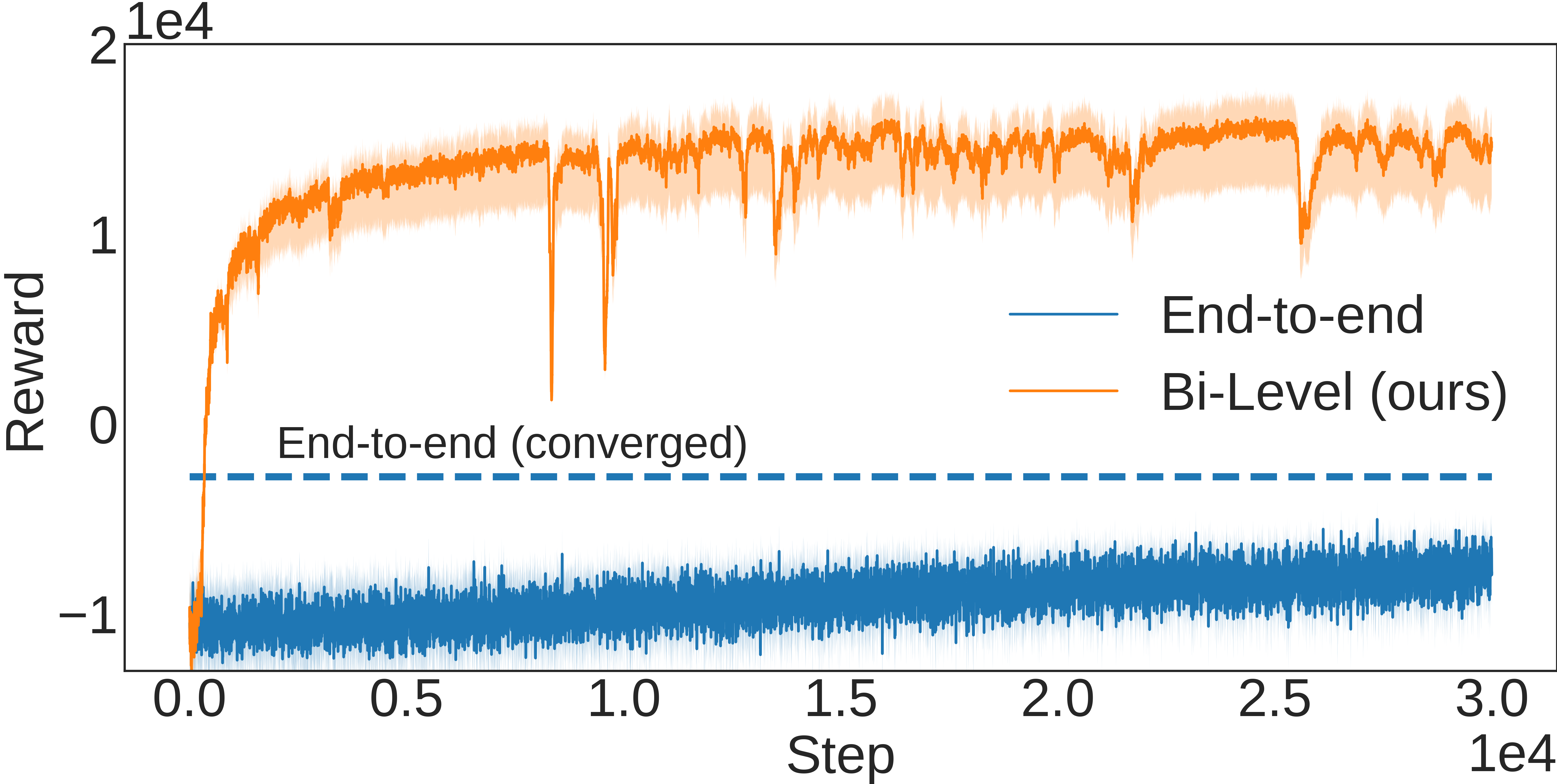}
      \caption{Learning curve comparison between an RL agent trained end-to-end (blue) and via our bi-level formulation (orange) on the DVR New York environment. The dotted line represents the converged performance for the end-to-end agent after 80,000 steps.}
      \label{fig:training_curve_dvr}
\end{figure}

Results in Figure \ref{fig:training_curve_dvr} highlight the sample efficiency of our bi-level approach compared to its end-to-end counterpart which exhibits (i) much slower convergence and sample inefficiency, and (ii) worse overall performance. 

\clearpage
\section{Additional Visualizations}

\end{document}